\PassOptionsToPackage{table,xcdraw}{xcolor}
\documentclass[10pt]{article} 
\usepackage[preprint]{tmlr}

\usepackage{amsmath,amsfonts,bm}

\def\eqref#1{equation~\ref{#1}}

\def\1{\bm{1}}

\DeclareMathAlphabet{\mathsfit}{\encodingdefault}{\sfdefault}{m}{sl}
\SetMathAlphabet{\mathsfit}{bold}{\encodingdefault}{\sfdefault}{bx}{n}

\usepackage{hyperref}
\usepackage{url}

\usepackage{amsmath}
\usepackage[linesnumbered,ruled,vlined]{algorithm2e}
\usepackage{multirow}
\usepackage{adjustbox}
\SetKwInput{kwInit}{Init}
\SetKwInput{KwRequire}{Require}
\usepackage{graphicx, caption, subcaption}
\usepackage{booktabs}

\definecolor{LightBlue}{rgb}{0.88,0.95,1.0} 
\newcommand{\best}[1]{\textbf{#1}}
\usepackage{booktabs}
\usepackage{multirow}
\usepackage{tabularx}
\usepackage{graphicx}
\usepackage{caption}
\usepackage{xspace}

\usepackage{adjustbox}       
\usepackage{colortbl}        
\usepackage{xcolor}
\usepackage{multirow}        
\usepackage{booktabs}        
\definecolor{LightBlue}{rgb}{0.8, 0.9, 1}

\newtheorem{theorem}{Theorem}[section]
\newtheorem{proposition}[theorem]{Proposition}

\newtheorem{definition}[theorem]{Definition}

\title{On the Role of Discrete Representation in Sparse Mixture of Experts}

\author{
        Giang Do\thanks{Corresponding author} \quad Kha Pham \quad Hung Le \quad Truyen Tran  \\
    Applied Artificial Intelligence Institute (A2I2), Deakin University \\
    \texttt{\{truong.do,phti, thai.le,truyen.tran\}@deakin.edu.au}\
      }

\def\month{MM}  
\def\year{YYYY} 
\def\openreview{\url{https://openreview.net/forum?id=XXXX}} 

\begin{document}

\maketitle

\begin{abstract}
  Sparse Mixture of Experts (SMoE) is an effective solution for scaling up model capacity without increasing the computational costs.
  A crucial component of SMoE is the router, responsible for directing the input to relevant experts; however, it also presents a major weakness, leading to routing inconsistencies and representation collapse issues.
  Instead of fixing the router like previous works, we propose an alternative that assigns experts to input via \emph{indirection}, which employs the discrete representation of input that points to the expert.
  The discrete representations are learned via vector quantization, resulting in a new architecture dubbed Vector-Quantized Mixture of Experts (VQMoE).
  We provide theoretical support and empirical evidence demonstrating the VQMoE's ability to overcome the challenges present in traditional routers.
  Through extensive evaluations on both large language models and vision tasks for pre-training and fine-tuning, we show that VQMoE achieves a 28\% improvement in robustness compared to other SMoE routing methods while maintaining strong performance in fine-tuning tasks.
\end{abstract}

\section{Introduction}
\label{intro}

Scaling Transformer models with increasing data and computational resources has led to remarkable advances across a wide range of domains, including natural language processing (NLP) \citep{du2022glam, fedus2022switch, zhou2024brainformers} and visual representation learning \citep{riquelme2021scalingvisionsparsemixture, shen-etal-2023-scaling}. Despite these successes, training and deploying large-scale dense Transformer models often require substantial computational resources, frequently amounting to hundreds of thousands of GPU hours and incurring costs in the millions of dollars \citep{kaddour2023challenges}. To address this scalability bottleneck, Sparse Mixture of Experts (SMoE) architectures have emerged as a promising alternative \citep{shazeer2017outrageously, zoph2022stmoe, xue2024openmoe, jiang2024mixtral}. Inspired by classical Mixture of Experts formulations \citep{6797059}, SMoE models consist of multiple expert subnetworks with shared architectures, where a routing mechanism dynamically selects a small subset of experts (often one or two) for each input token. This sparsity significantly reduces inference costs compared to dense counterparts of similar model capacity \citep{artetxe2022efficient, krajewski2024scaling}, making SMoEs attractive for efficient scaling.

Despite their efficiency benefits, SMoEs face critical training challenges, most notably, \emph{representation collapse}. This phenomenon occurs when only a small subset of experts are frequently activated, or when all experts converge to similar representations, thereby negating the diversity and specialization that the architecture is intended to promote. Prior works have sought to mitigate this issue by improving the routing policy through regularization and auxiliary losses \citep{chi2022representation, chen2023sparse, do2023hyperrouter}. However, these approaches focus on the routers improvement rather than questioning its necessity.

In this work, we explore a more fundamental question: \emph{Is an explicit router necessary at all?} We argue that incorporating discrete representations offers a principled alternative. Discrete latent variables are inherently suited to capturing structured and interpretable patterns within data, aligning with the symbolic nature of human cognition, where concepts are often discretized as words, tokens, or categories. In the SMoE context, discrete representations can improve input routing by naturally clustering similar inputs, thereby enhancing expert specialization and utilization without relying solely on a learned gating mechanism.


Employing vector quantization (VQ) techniques to learn discrete representation, this paper proposes a novel mixture of expert framework, named VQMoE, which overcomes the representation collapse and inconsistency in training sparse mixture of experts. More specifically, we prove that the existing router methods are inconsistent and VQMoE suggests an optimal expert selection for training SMoE. Additionally, our method guarantees superior SMoE training strategies compared to the existing methods by solving the representation collapse by design.

We evaluate the proposed method by conducting pre-training of Large Language Models (LLMs) on several advanced SMoE architectures, such as SMoE~\citep{jiang2024mixtral}, StableMoE~\citep{dai2022stablemoe}, or XMoE~\citep{chi2022representation}, followed by fine-tuning on downstream tasks on both Language and Vision domains.

In summary, the primary contributions of this paper are as follows:
\begin{itemize}
    \item We theoretically demonstrate that learning discrete representations provides an effective mechanism for expert selection, and that VQMoE intrinsically mitigates the problem of representation collapse.
    \item We propose the use of vector quantization (VQ) to learn structured and interpretable expert clusters.
    \item We conduct extensive experiments on large language models as well as vision pre-training and fine-tuning tasks to validate the effectiveness of our method.
    \item We provide a comprehensive analysis of VQMoE’s behavior, offering insights into its performance and robustness.
\end{itemize}

\section{Related Work}\label{sec:related}

\textbf{Sparse Mixture of Experts (SMoE).} Sparse Mixture of Experts (SMoE) builds on the Mixture of Experts (MoE) framework introduced by \citet{jacobs1991,jordan1994}, with the core idea that only a subset of parameters is utilized to process each example. This approach was first popularized by \citet{shazeer2017outrageously}. SMoE's popularity surged when it was combined with large language models based on Transformers~\citep{NEURIPS2022_2f00ecd7,li2022branchtrainmerge,shen2023mixtureofexperts}, and its success in natural language processing led to its application across various fields, such as computer vision~\citep{NEURIPS2021_48237d9f,hwang2023tutel,lin2024moellava}, speech recognition~\citep{wang2023languagerouting,kwon2023}, and multi-task learning~\citep{Ye_2023_ICCV,Chen_2023_CVPR}.

However, SMoE faces a major problem in training known as representation collapse, i.e., the experts converge to similar outputs. To address this, various methods have been introduced. XMoE~\citep{chi2022representation} calculates routing scores between tokens and experts on a low-dimensional hypersphere.
SMoE-dropout~\citep{chen2023sparse} uses a fixed, randomly initialized router network to activate experts and gradually increase the number of experts involved to mitigate collapse. Similarly, HyperRouter~\citep{do2023hyperrouter} utilizes HyperNetworks~\citep{ha2016hypernetworks} to generate router weights, providing another pathway for training SMoE effectively. StableMoE~\citep{dai2022stablemoe} introduces a balanced routing approach where a lightweight router, decoupled from the backbone model, is distilled to manage token-to-expert assignments. The StableMoE strategy ensures stable routing by freezing the assignments during training, while SimSMoE~\cite{do2024simsmoesolvingrepresentationalcollapse} forces experts to learn dissimilar representations. Despite these extensive efforts, the representation collapse issue persists, as highlighted by \citet{pham2024competesmoe}. While most solutions focus on improving routing algorithms, our approach takes a different path by learning a discrete representation of input that points to relevant experts.

\textbf{Discrete Representation.}\; Discrete representations align well with human thought processes; for example, language can be understood as a series of distinct symbols. 
Nevertheless, the use of discrete variables in deep learning has proven challenging, as evidenced by the widespread preference for continuous latent variables in most current research. VQVAE~\citep{NIPS2017_7a98af17} implements discrete representation in Variational AutoEncoder (VAE)~\citep{kingma2022autoencodingvariationalbayes} using vector quantization (VQ).  IMSAT~\citep{pmlr-v70-hu17b} attains a discrete representation by maximizing the information-theoretic dependency between data and their predicted discrete representations. Recent works follow up the vector quantization ideas and make some enhancements for VAE, for example:~\citep{yu2022vectorquantizedimagemodelingimproved};  ~\citep{mentzer2023finitescalarquantizationvqvae}; and \citep{yang2023hificodecgroupresidualvectorquantization}. \citet{mao2022discreterepresentationsstrengthenvision} utilize a discrete representation to strengthen Vision Transformer (ViT)~\citep{dosovitskiy2021imageworth16x16words}. To the best of our knowledge, our paper is the first to learn a discrete representation of Sparse Mixture of Experts. 

\vspace{-0.1in}
\section{Method}
\label{method}

We propose a novel model, Vector-Quantized Mixture of Experts (VQMoE), which learns discrete representations for expert selection. As illustrated in Fig.~\ref{fig:old}, our approach selects experts directly based on the input representation, eliminating the need for a trained router. To prevent information loss, we integrate discrete and continuous representations within the model.

\vspace{-0.1in}
\subsection{Preliminaries}





\textbf{Sparse Mixture of Experts.}\; Sparse Mixture of Experts (SMoE) is a variant of the transformer architecture in which the conventional feed-forward layers (MLPs) are replaced with Mixture of Experts (MoE) layers~\citep{shazeer2017outrageously}. 
Given an input $\boldsymbol{x} \in \mathbb{R}^{n \times d}$, which represents the output of the multi-head attention (MHA) module, the SMoE layer computes a sparse weighted combination over a set of $N$ expert networks. 
Each expert is typically a feed-forward neural network $FFN_i(\boldsymbol{x})$, and its contribution to the final output is determined by a routing function $\mathcal{S}(\boldsymbol{x})$. 
The resulting output of the SMoE layer is given by:

\begin{equation}
\begin{aligned}
    f^{\mathrm{SMoE}}(\boldsymbol{x}) &= \sum_{i=1}^N \mathcal{S}(\boldsymbol{x})_i \cdot FFN_i(\boldsymbol{x}) \\
    &= \sum_{i=1}^N \mathcal{S}(\boldsymbol{x})_i \cdot \boldsymbol{W}_{\mathrm{FFN}_i}^2
    \, \phi\left(\boldsymbol{W}_{\mathrm{FFN}_i}^1 \boldsymbol{x}\right),
\end{aligned}
\label{eq:smoe}
\end{equation}

\noindent
where $\phi(\cdot)$ denotes a non-linear activation function (e.g., ReLU or GELU), and $\boldsymbol{W}_{\mathrm{FFN}_i}^1$, $\boldsymbol{W}_{\mathrm{FFN}_i}^2$ are the learnable weights of the $i$-th expert. 
The routing weights $\mathcal{S}(\boldsymbol{x})$ are computed using a Top-$k$ selection over the softmax scores derived from the dot product of the input with a learned expert embedding matrix $\boldsymbol{W}_e$, as defined below:

\begin{equation}
\begin{aligned}
    \mathcal{S}(\boldsymbol{x}) &= \operatorname{TopK}(\operatorname{softmax}(\boldsymbol{W}_e \boldsymbol{x}), k), \\
    \operatorname{TopK}(\boldsymbol{v}, k) &= 
    \begin{cases} 
        \boldsymbol{v}_i & \text{if } \boldsymbol{v}_i \in \text{top } k \text{ largest elements of } \boldsymbol{v}, \\
        -\infty & \text{otherwise}.
    \end{cases}
\end{aligned}
\label{eq:topk}
\end{equation}

\noindent
This sparse selection mechanism ensures that only a small subset of experts are activated for each input, which significantly reduces computational cost while retaining model capacity.

\textbf{Discrete Representation Learning.}\;  \citet{NIPS2017_7a98af17} propose VQVAE, which uses Vector Quantization (VQ) to learn a discrete representation. Given an input $x \in \mathbb{R}^{n \times d}$, VQVAE discretized the input into a codebook $V \in \mathbb{R}^{K \times d}$ where $K$ is the codebook size and $d$ is the dimension of the embedding. Let denote $z_v(x) \in \mathbb{R}^{n \times d}$ the output of the VQVAE and $\mathbf{1}()$ is the indicator function. The discrete representation $z_q(x_i)=v_k, \quad \text { where } \quad k=\operatorname{argmin}_j\left\|z_v(x_i)-v_j\right\|_2$ is achieved by vector quantizer $q_\theta$ that maps an integer $z$ for each input $x$ as: 
\begin{equation}
q_\theta(z=k \mid x)=\mathbf{1}\left(k=\underset{j=1: K}{\arg \min }\left\|z_v(x)-\mathrm{V}_j\right\|_2\right)
\end{equation}


\subsection{Vector-Quantized Mixture of Experts (VQMoE)}
\label{arch}

\begin{figure*}[t]
    \centering
    \begin{subfigure}{.54\textwidth}
         \centering
         \includegraphics[width=\textwidth]{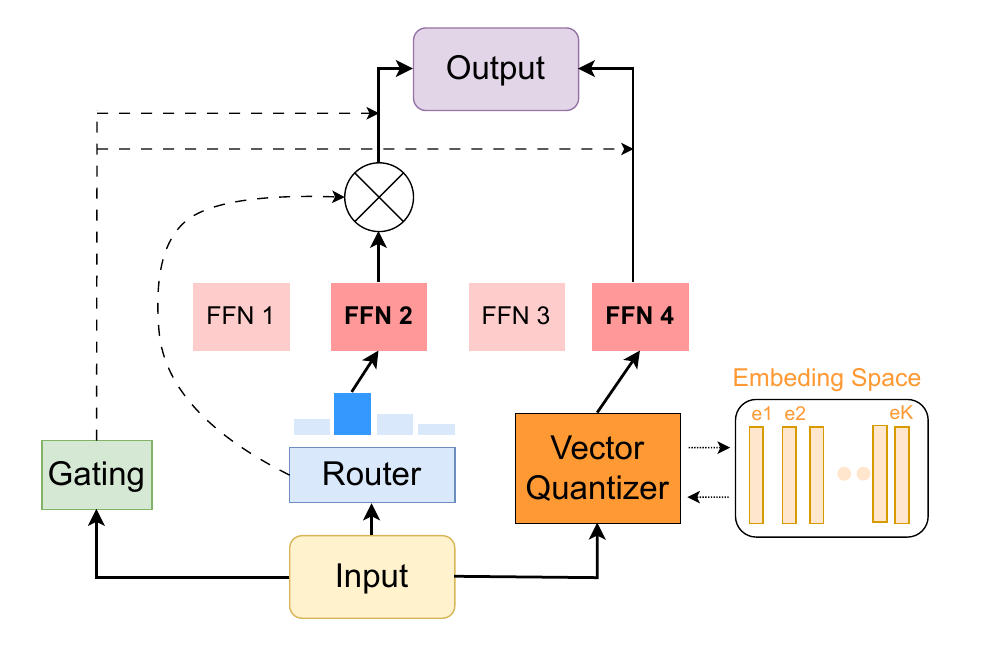}
         \caption{VQMoE Pre-training }
         \label{fig:old}
     \end{subfigure}
     \hfill
    \begin{subfigure}{.42\textwidth}
         \centering
         \includegraphics[width=\textwidth]{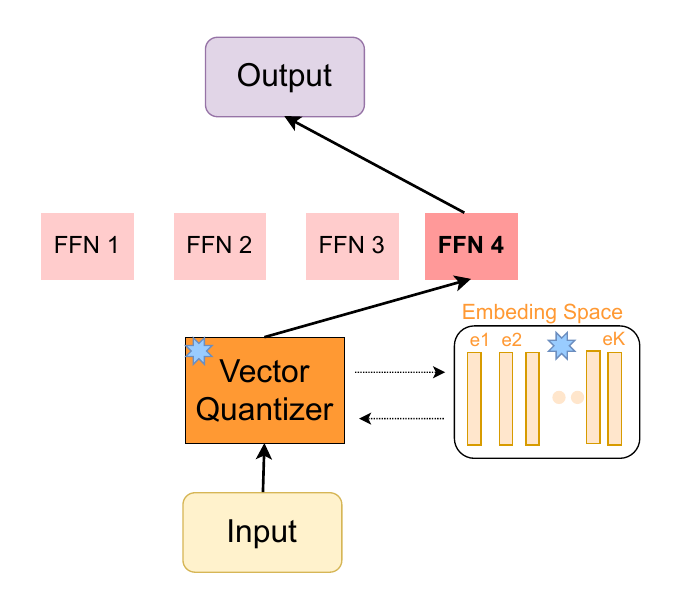}
         \caption{VQMoE Fine-tuning }
         \label{fig:new}
     \end{subfigure}
     \hfill
     
     \caption{Illustration of the proposed VQMoE architecture for Pre-training and fine-tuning. (a) At the Pre-training stage, VQMoE architecture learns simultaneously continuous and discrete representation at the Pre-training phase. The continuous representation is learned by the conventional SMoE, while the Vector Quantization block facilitates the learning of a discrete representation. The final output is then combined by a gate layer. (b) VQMoE learns a discrete representation that is capable of operating efficiently and robustly on downstream tasks. VQMoE computes the discrete representation only during the fine-tuning stage to achieve robustness and efficiency.} \label{fig:simsmoe}
     \vspace{-0.1in}
\end{figure*}

\textbf{Pre-training VQMoE.}\;
Traditional Sparse Mixture of Experts (SMoE) models utilize continuous token representations and route them to experts based on learned token-expert affinity scores. We propose a novel architecture, VQMoE, that learns both continuous and discrete representations jointly during pre-training (see Figure~\ref{fig:old}). The continuous component captures fine-grained data patterns, while the discrete component, learned via vector quantization, encodes robust latent structure useful for downstream transfer.

Let $\boldsymbol{x} \in \mathbb{R}^{n \times d}$ denote the input to the VQMoE layer (e.g., output from a multi-head attention block), and let $f^{\mathrm{vq}}$ denote the vector quantization operator. The VQMoE output during pre-training is defined as:

\begin{equation}
f^{\mathrm{VQMoE}}(\boldsymbol{x}) = \underbrace{g_c(\boldsymbol{x}) \cdot f^{\mathrm{SMoE}}(\boldsymbol{x})} + \underbrace{g_d(\boldsymbol{x}) \cdot \sum_{l=1}^K f_l^{\mathrm{FFN}}(\tilde{\boldsymbol{x}}_l)}
\end{equation}
\vspace{-6mm}
\begin{center}
\hspace{15mm} \textit{(Continuous representation)} \hspace{5mm} \textit{(Discrete representation)}
\end{center}
\vspace{1mm}


    
    
    

In this formulation, $f^{\mathrm{SMoE}}(\boldsymbol{x})$ denotes the output from a standard Sparse Mixture of Experts (SMoE) layer, capturing the continuous expert representations. The second term corresponds to the discrete representation, where each $f_l^{\mathrm{FFN}}$ is the $l$-th feedforward expert network. The input to each discrete expert, denoted as $\tilde{\boldsymbol{x}}_l$, is determined by vector quantization: specifically, $\tilde{\boldsymbol{x}}_l = \boldsymbol{v}_k$ if the input vector $\boldsymbol{x}_l$ is assigned to the $l$-th codebook vector $\boldsymbol{v}_k$; otherwise, it is set to the zero vector, i.e., $\tilde{\boldsymbol{x}}_l = \boldsymbol{0}$. Here, $K$ is the number of vector quantization codebooks, and $\boldsymbol{v}_k$ is a learned codebook vector assigned by $f^{\mathrm{vq}}$. The gating functions $g_c(\boldsymbol{x})$ and $g_d(\boldsymbol{x})$ as Equation ~\ref{eqa_gating_22}, modulate the contributions of the continuous and discrete pathways, respectively, and are typically computed based on the input $\boldsymbol{x}$ through learnable mechanisms.

\begin{equation}\label{eqa_gating_22}
\begin{bmatrix}
g_c(\boldsymbol{x}) \
g_d(\boldsymbol{x})
\end{bmatrix}
= \operatorname{softmax}(W_g \boldsymbol{x}), \quad W_g \in \mathbb{R}^{2 \times d}
\end{equation} 







To address the mismatch between the number of codebook vectors and the number of expert networks, we introduce a \textit{flexible code} strategy. This approach enables consistent routing from quantized representations to experts, even when the two quantities differ. Specifically, we define a hash-based mapping using a modulo operation. Let \( i_{\mathrm{cb}} \) denote the index of a codebook vector, and let \( i_{\mathrm{exp}} \) denote the index of the corresponding expert. The mapping is given by:

\begin{equation}
    i_{\mathrm{exp}} = i_{\mathrm{cb}} \bmod N,
\end{equation}

\noindent
where \( N \) is the total number of experts. This ensures each codebook index is deterministically assigned to one of the available experts

\textbf{Fine-tuning VQMoE.}\; Based on insights from \citet{geva-etal-2021-transformer}, which note that feed-forward layers (FFNs) constitute a significant portion of a transformer's parameters, we adopt a lightweight fine-tuning strategy that retains only the discrete path of the VQMoE. This allows efficient adaptation while leveraging pre-trained latent representations (see Figure~\ref{fig:new}). The fine-tuning output becomes:

\begin{equation}
f^{\mathrm{VQMoE}}(\boldsymbol{x}) = \sum_{l=1}^K f_l^{\mathrm{FFN}}(\tilde{\boldsymbol{x}}_l)
\end{equation}

\vspace{-0.1in}
\subsection{Training Procedure}
\label{training}
\textbf{Pretraining.}\; The training objective is jointly minimizing the loss of the target task and losses of the Vector Quantization module ($\mathcal{L}^{\text {l2 }}$ and $\mathcal{L}^{\text {commitment }}$) as in \citep{NIPS2017_7a98af17}. Equation \ref{eqa:loss} specifies the overall loss function for training VQMoE with three components: (1) task loss; (2) $l_2$ loss; (3) a commitment loss. While $\mathcal{L}^{\text {l2 }}$ helps to move the embedding $v_i$ towards the outputs $z_v(x)$, the commitment loss makes sure the output of the Vector Quantization module commits to the embedding and its output does not grow. The Vector Quantization algorithm does not vary with $\beta$, we follow $\beta=0.25$ as \citet{NIPS2017_7a98af17}. We introduce a new parameter, $\alpha$, to regulate the contribution of the Vector Quantization loss to the overall loss. A higher value of $\alpha$ favors a stronger adherence to the discrete representation, and vice versa. 

\begin{equation} \label{eqa:loss}
L=\mathcal{L}^{\text {task }}+\alpha( \left\|\operatorname{sg}\left[z_v(x)\right]-v\right\|_2^2+ \beta\left\|z_v(x)-\operatorname{sg}[v]\right\|_2^2)
\end{equation}
where $sg(.)$ is the stop gradient operator defined as follows: 
\begin{equation}
\operatorname{sg}(x)= \begin{cases}x & \text { forward pass } \\ 0 & \text { backward pass }\end{cases}
\end{equation}
In Equation~\ref{eqa:loss}, \( \mathcal{L}^{\text{task}} \) denotes the task-specific loss, which depends on applications. For example, in language modeling tasks, \( \mathcal{L}^{\text{task}} \) is typically defined as the negative log-likelihood (NLL) of the target tokens~\citep{dai2019transformerxlattentivelanguagemodels}, promoting accurate next-token prediction. In image classification tasks, \( \mathcal{L}^{\text{task}} \) is usually implemented as the cross-entropy loss between the predicted class distribution and the ground-truth label~\citep{he2015deepresiduallearningimage}, encouraging correct class assignment.

\textbf{Fine-tuning.}\; For downstream tasks, we fine-tune the pretraining model by utilizing the codebook learned from the Equation ~\ref{eqa:loss} by freezing all parameters at the Vector Quantization module. Thus, the training objective simply becomes: $L=\mathcal{L}^{\text {task }}$. 

\section{Theory Analysis}
\label{sec:theory}

\subsection{Optimal Experts Selection}
\textbf{Problem settings.}\; We consider an MoE layer with each expert being an MLP layer which is trained by gradient descent and input data $\left\{\left(\mathbf{x}_i, y_i\right)\right\}_{i=1}^n$ generated from a data distribution $\mathcal{D}$. Same as ~\citep{NEURIPS2022_91edff07}; ~\citep{dikkala-etal-2023-benefits}, we assume that the MoE input exhibits cluster properties, meaning the data is generated from $N$ distinct clusters $(C_1, C_2, ..., C_N)$.  

\begin{definition}[Consistent Router]
    \label{definition:consistent}
    A sequence of points \( x_1, x_2, \ldots, x_n \) and a corresponding sequence of clusters \( C_1, C_2, \ldots, C_N \) are said to be \textbf{consistent} if, for every point \( x_p \in C_i \), the condition  
    \[
    \text{dist}(x_p, u_i) \leq \min_{j \neq i} \text{dist}(x_p, u_j)
    \]  
    is satisfied, where \( \text{dist}(a, b) \) denotes the distance between \( a \) and \( b \), and \( u_i \) is the center of cluster \( C_i \).
\end{definition}



\begin{definition}[Inconsistent Router]
\label{definition:inconsistent}  
A sequence of points \( x_1, x_2, \ldots, x_n \) and a corresponding sequence of clusters \( C_1, C_2, \ldots, C_N \) are said to be \textbf{inconsistent} if there exists a point \( x_p \in C_i \) such that  
\[
\text{dist}(x_p, u_i) > \min_{j \neq i} \text{dist}(x_p, u_j),
\]  
where \( \text{dist}(a, b) \) represents the distance between \( a \) and \( b \), and \( u_i \) is the center of cluster \( C_i \).
\end{definition}


Inspired by ~\citep{dikkala-etal-2023-benefits}, we conceptualize the router in Sparse Mixture of Experts as a clustering problem. This leads us to define a consistent router in Definition ~\ref{definition:consistent}. Furthermore, we introduce a definition for an inconsistent router in SMoE as outlined in Definition ~\ref{definition:inconsistent}, along with the concept of inconsistent expert selection presented in Theorem \ref{theorem:inconsistent_select} during the training of SMoE.


\begin{theorem}[Inconsistent Experts Selection]
    \label{theorem:inconsistent_select}
    Let \( f_{MHA} \) be a multi-head attention (MHA) function producing an output \( x \in \mathbb{R}^{n \times d} \), and consider \( N \) experts with embeddings \( e_i \) for expert \( i \) where \( i \in [1, N] \). Assume that \( f_{MHA} \) converges at step \( t_m \), while the expert embeddings \( e \) converge at step \( t_e \), with \( t_m \gg t_e \). For each output \( x \), an expert \( P \in [1, N] \) is selected such that 
    \[
    P = \arg\min_{j \in [1, N]} \text{dist}(x, e_j).
    \]
    Under these conditions, the expert embeddings \( e \) form an inconsistent routing mechanism.
\end{theorem}

The proof of Theorem ~\ref{theorem:inconsistent_select} is given in Appendix ~\ref{appendix:inconsistent_select}, and we have the following insights. Theorem ~\ref{theorem:inconsistent_select} implies that an expert selection process by a router as the conventional SMoE leads to the inconsistent router. Indeed, the router layer is designed as a simple linear layer, $x$ is the output of MHA function in practice; and an SMoE router is significantly simpler than the MHA function. Consequently, this design leads to the router functioning as an inconsistent router, contributing to the representation collapse issue and instability during training.


\begin{proposition}[Optimal Experts Selection]
    \label{prop:exp_discrete}
    Given input data partitioned into \( N \) clusters \( (C_1, C_2, \ldots, C_N) \) and a mixture of experts (MoE) layer with \( N \) experts \( (E_1, E_2, \ldots, E_N) \), the assignment of each cluster \( C_i \) to expert \( E_i \) for \( i \in [1, k] \) constitutes an optimal expert selection solution.
\end{proposition}

Proposition ~\ref{prop:exp_discrete} demonstrates that if we are given a clustering structure as input, assigning each part of the input to its corresponding expert results in an optimal expert selection. This implies that learning a discrete representation and directing each component to the appropriate expert yields an optimal solution. The proof of Proposition ~\ref{prop:exp_discrete} can be found in Appendix ~\ref{appendix:optimal}.

\subsection{Experts Representation Collapse}

The \textit{representation collapse} problem in Sparse Mixture of Experts (SMoE), where all experts converge to similar representations, was first highlighted by \citet{chi2022representation}. Following \citet{chi2022representation} and \citet{do2023hyperrouter}, we analyze this issue using the Jacobian matrix of the model output with respect to the input \( x \in \mathbb{R}^{n \times d} \). The Jacobian for SMoE is expressed as:

\begin{equation}
\label{eq:jacobian_smoe}
\begin{aligned}
\boldsymbol{J}^{\text{SMoE}} &= \mathcal{S}(x)_k \boldsymbol{J}^{\text{FFN}} 
+ \sum_{j=1}^N \mathcal{S}(x)_k(\delta_{kj} - \mathcal{S}(x)_j) \boldsymbol{E}(x)_i \boldsymbol{e}_j^\top \\
&= \mathcal{S}(x)_k \boldsymbol{J}^{\text{FFN}} + \sum_{j=1}^N \boldsymbol{c}_j \boldsymbol{e}_j^\top,
\end{aligned}
\end{equation}

where \( \boldsymbol{c}_j = \mathcal{S}(x)_k(\delta_{kj} - \mathcal{S}(x)_j) \boldsymbol{E}(x)_i \), \( \boldsymbol{J}^{\text{FFN}} \) is the Jacobian of the selected expert's feedforward network, and \( \boldsymbol{e}_j \) are the expert embedding vectors. Equation~\ref{eq:jacobian_smoe} consists of two components: \( \mathcal{S}(x)_k \boldsymbol{J}^{\text{FFN}} \) - the main signal path from the input to the output through the selected expert; and \( \sum_{j=1}^N \boldsymbol{c}_j \boldsymbol{e}_j^\top \) - the contribution from the gating function's gradient with respect to the expert embeddings.


Since the summation over expert embeddings lies in a subspace of dimension \( N \), and typically \( N \ll d \), this projection restricts the output space from \( \mathbb{R}^d \) to \( \mathbb{R}^N \), which effectively causes representation collapse.


\textbf{Jacobian Analysis of VQMoE.}\; To examine whether VQMoE mitigates this collapse, we derive the Jacobian of the VQMoE output with respect to the input \( x \in \mathbb{R}^{n \times d} \). The detailed expression of the VQMoE Jacobian matrix is provided in Section~\ref{app:vqproof}. Specifically, we have: 




\begin{equation}
\label{eq:jacobian_vqmoe}
\begin{aligned}
\boldsymbol{J}^{\text{VQMoE}} 
&= g_c(\boldsymbol{x}) \cdot \boldsymbol{J}^{\text{SMoE}} 
+ \frac{\partial g_c(\boldsymbol{x})}{\partial \boldsymbol{x}} f^{\mathrm{SMoE}}(\boldsymbol{x}) \\
&\quad + g_d(\boldsymbol{x}) \cdot \sum_{l=1}^K \boldsymbol{J}^{\text{FFN}}_l 
+ \frac{\partial g_d(\boldsymbol{x})}{\partial \boldsymbol{x}} \sum_{l=1}^K f_l^{\mathrm{FFN}}(\tilde{\boldsymbol{x}}_l) \\
&= J_1 + \sum_{j=1}^{N+K+2} o_j \boldsymbol{e}_j^\top.
\end{aligned}
\end{equation}

Same as the Jacobian matrix of SMoE, the Jacobian matrix of VQMoE consists two terms: (1) $J_1$ depends on input token and experts to the final output; (2) $\sum_{j=1}^{N+K+2} o_j \boldsymbol{e}_j^{\top}$ indicates to learn better gating function to minimize the task loss. We can see that $N+K+2 >> N$, implying that VQMoE is better than SMoE in solving the representation collapse issue. In theory, we can choose the number of codebook to be approximately $d-N-2$ with a hashing index to experts to address the issue. However, this involves a trade-off with the computational resources required to learn the codebook.

\vspace{-0.1in}
\section{Experiment} \label{sec:exp}
\vspace{-0.1in}

We conduct experiments to investigate the following hypotheses: (i) VQMoE offers an effective training algorithm for Sparse Mixture-of-Experts (SMoE) in large language models (LLMs); (ii) VQMoE enables efficient fine-tuning; and (iii) VQMoE outperforms other routing methods across multiple domains.

\subsection{Experimental Settings}

To evaluate the three hypotheses, we conduct experiments across both vision and language tasks. For pre-training language models, we assess two standard benchmarks: (i) character-level language modeling using enwik8 and text8\citep{mahoney_large_2011}, and (ii) word-level language modeling using WikiText-103\citep{merity2016pointersentinelmixturemodels} and the more challenging One Billion Word (lm1b) dataset~\citep{chelba2014billionwordbenchmarkmeasuring}. All experiments use the standard training, validation, and test split with a 90:5:5 ratio as\citep{child2019generatinglongsequencessparse}.

For parameter-efficient fine-tuning, we fine-tune models pre-trained on enwik8 using four widely used NLP datasets: SST-2, SST-5\citep{socher_recursive_2013}, IMDB\citep{maas_learning_2011}, and BANKING77\citep{casanueva-etal-2020-efficient}. Following\citet{chen2023sparse}, we freeze the router and update only the expert parameters to evaluate fine-tuning efficiency.

For vision tasks, we employ the Vision Transformer (ViT)\citep{dosovitskiy2021imageworth16x16words} and compare our routing method with state-of-the-art alternatives on five benchmark image classification datasets: CIFAR-10, CIFAR-100\citep{Krizhevsky09learningmultiple}, STL-10\citep{coates2011stl10}, SVHN\citep{Netzer2011}, and ImageNet-1K~\citep{deng2009imagenet}.

\subsection{Pre-training Language Models}\label{sec:pretrained}

\begin{table*}[t]
\centering
\begin{adjustbox}{width=1\textwidth}
\begin{tabular}{@{}llcccccccc>{\columncolor{LightBlue}}c>{\columncolor{LightBlue}}c@{}}
\toprule
\multicolumn{2}{c}{Configuration}                & \multicolumn{2}{c}{Enwik8 (BPC)} & \multicolumn{2}{c}{Text8 (BPC)} & \multicolumn{2}{c}{WikiText-103 (PPL)} & \multicolumn{2}{c}{lm1b (PPL)} & Avg. Char-level & Avg.Word-level \\ \midrule
Architecture                 & Algorithm   & Base & Large  & Base & Large & Base & Large & Base & Large & - & - \\ \midrule

\multirow{5}{*}{Transformer} 
& VQMoE         & \textbf{1.48} & \textbf{1.41} & \textbf{1.47} & \textbf{1.40} & \textbf{38.74} & \textbf{31.98} & \textbf{59.48} & \textbf{49.30} & \textbf{1.44} & \textbf{44.88} \\
& SMoE          & 1.49 & \textbf{1.41} & 1.49 & \textbf{1.40} & 39.50 & 32.30 & 60.88 & 51.30 & 1.45 & 45.50 \\
& SMoE-Dropout  & 1.82 & 2.22 & 1.70 & 1.89 & 72.62 & 107.18 & 97.45 & 159.09 & 1.91 & 109.59 \\
& XMoE          & 1.51 & 1.42 & 1.49 & 1.42 & 39.56 & 32.65 & 61.17 & 51.84 & 1.46 & 46.06 \\
& StableMoE     & 1.49 & 1.42 & 1.49 & 1.41 & 39.45 & 32.34 & 60.72 & 50.74 & 1.45 & 45.81 \\ \midrule

\multirow{5}{*}{Transformer-XL} 
& VQMoE         & \textbf{1.19} & \textbf{1.08} & \textbf{1.28} & \textbf{1.17} & \textbf{29.48} & \textbf{23.85} & \textbf{56.85} & \textbf{48.70} & \textbf{1.18} & \textbf{39.72} \\
& SMoE          & 1.20 & 1.09 & 1.29 & 1.18 & 30.16 & 24.02 & 58.00 & 48.71 & 1.19 & 40.22 \\
& SMoE-Dropout  & 1.56 & 2.24 & 1.56 & 1.86 & 58.37 & 40.02 & 93.17 & 68.65 & 1.81 & 65.55 \\
& XMoE          & 1.21 & 1.09 & \textbf{1.28} & \textbf{1.17} & 30.34 & 24.22 & 58.33 & 50.64 & 1.19 & 40.88 \\
& StableMoE     & 1.20 & 1.10 & \textbf{1.28} & 1.19 & 29.97 & 24.19 & 58.25 & 49.17 & 1.19 & 40.40 \\ \midrule

\multicolumn{2}{c}{$\#$ Params} & 20M & 210M & 20M & 210M & 20M & 210M & 20M & 210M & - & - \\ 
\bottomrule
\end{tabular}
\end{adjustbox}
\caption{Bits-per-character (BPC) on the Enwik8 and Text8 test sets; and perplexity (PPL) on the WikiText-103 and One Billion Word test sets. \textbf{Avg. Char-level} is the average BPC over Enwik8 and Text8; \textbf{Avg. Word-level} is the average PPL over WikiText-103 and lm1b. Lower is better; best results are in bold.}
\label{table:pre-train}
\end{table*}

\textbf{Models.}
For the language tasks, we follow the same settings as in SMoE-Dropout~\citep{chen2023sparse}. We consider two decoder-only architectures: (i) the standard Transformer~\citep{Vaswani+2017}; and (ii) and {Transformer-XL}~\citep{dai-etal-2019-transformer} with the same number of parameters as Transformer. We evaluate our method versus the state of art Sparse Mixture of Expert Layers such as StableMoE~\citep{dai2022stablemoe} and XMoE~\citep{chi2022representation} with top$k=2$ in the experiments. We consider two model configurations: (i) base: with four SMoE blocks and \textbf{20M} parameters; (ii) large: with twelve SMoE layers and \textbf{210M} parameters. We emphasize that we are not trying to achieve state-of-the-art results due to the limited resource constraints. Instead, we evaluate the small and large models on various datasets to demonstrate the scalability and efficacy of our algorithm. Lastly, we conduct extensive investigations using the tiny model to understand the algorithm behaviours and their robustness to different design choices. 


\textbf{Baselines.}
We compare our VQMoE with state-of-the-art SMoE training strategies for LLMs. \textbf{SMoE}~\citep{jiang2024mixtral} employs a simple router trained end-to-end with the experts. \textbf{StableMoE}~\citep{dai2022stablemoe} proposes a two-phase training process where the first phase trains only the router, and then the router is fixed to train the experts in the second phase. \textbf{XMoE}~\citep{chi2022representation} implements a deep router that comprises a down-projection and normalization layer and a gating network with learnable temperatures. Lastly, motivated by {SMoE-Dropout}~\citep{chen2023sparse}, we implement the \textbf{SMoE-Dropout} strategy that employs a randomly initialized router and freeze it throughout the training process.

\textbf{Training procedure.}\; 
For the language modeling experiments, we optimize the base models and the large models for 100,000 steps. We use an Adam~\citep{kingma2017adam} optimizer with a Cosine Annealing learning rate schedule~\citep{loshchilov2017sgdrstochasticgradientdescent}. The lowest validation loss checkpoint is used to report the final performance on the test set. 

\begin{figure*}[t]
    \centering
    \begin{subfigure}{.49\textwidth}
         \centering
         \includegraphics[width=\textwidth]{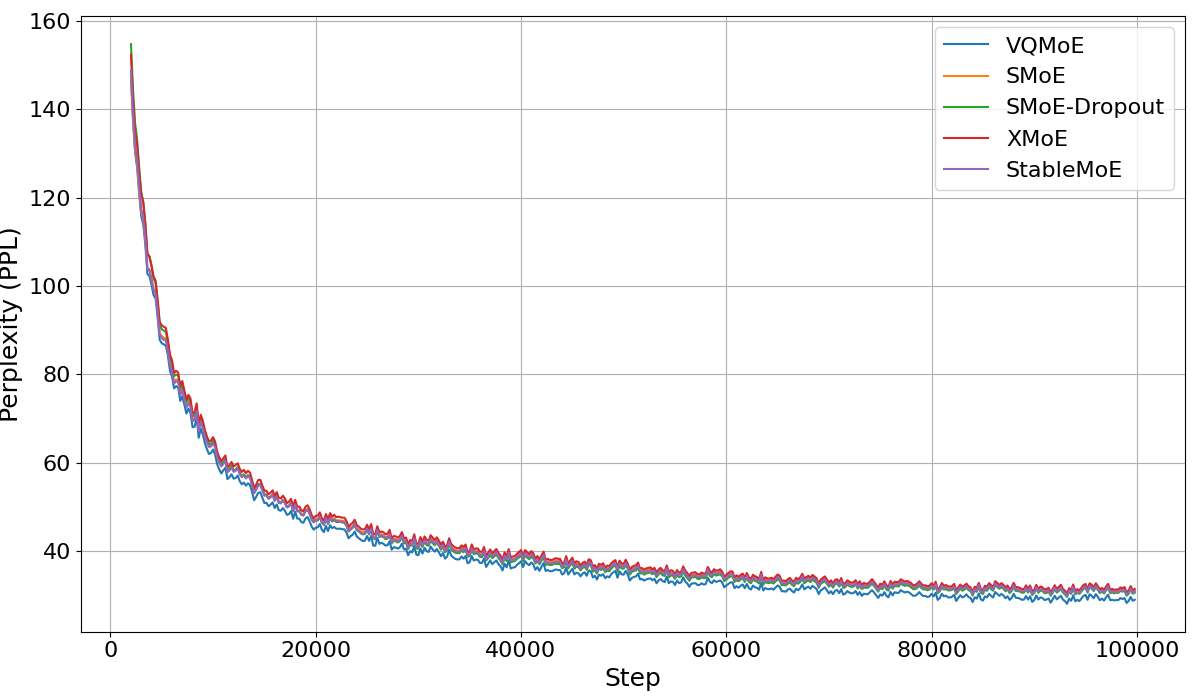}
         \caption{Training PPL movement on Wikitext-103 dataset.}
         \label{fig:wikitext103}
     \end{subfigure}
     \hfill
    \begin{subfigure}{.49\textwidth}
         \centering
         \includegraphics[width=\textwidth]{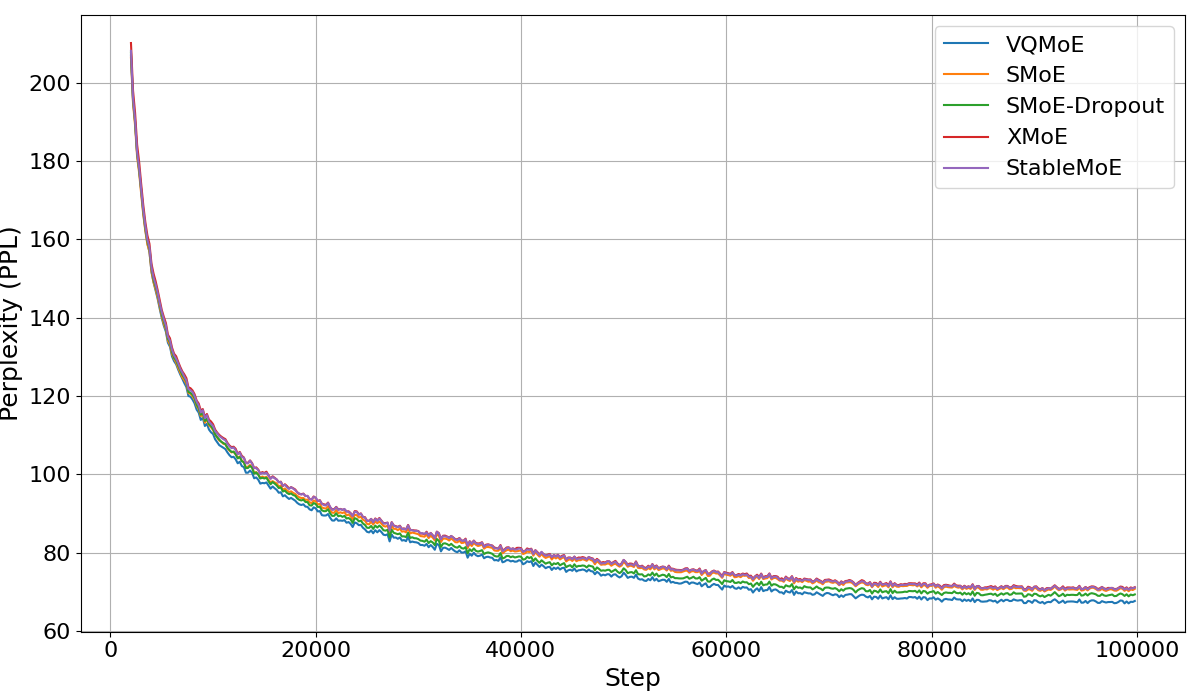}
         \caption{Training PPL movement on lm1b dataset.}
         \label{fig:1b}
     \end{subfigure}
     \hfill
     
     \caption{Perplexity (PPL) over training steps for the Transformer-XL base model on two datasets: (a) WikiText-103 and (b) lm1b. The results indicate that VQMoE converges faster than the baseline models, demonstrating its efficiency and robustness for language modeling tasks.} \label{fig:trainingloss}
     \vspace{-0.1in}
\end{figure*}

\textbf{\textit{Q1: Does VQMoE perform better on Pre-training tasks compared to routing methods? A1: Yes.}}

Table~\ref{table:pre-train} presents the evaluation metrics comparing VQMoE with state-of-the-art approaches. We also show the performance progression of the base model on the validation set. Notably, across all methods and datasets, \textbf{VQMoE consistently outperforms the baseline models} for both the Transformer-XL and Transformer architectures on average. Although advanced strategies such as XMoE and StableMoE generally outperform the vanilla SMoE on character-based datasets such as \textit{enwik8} and \textit{text8}, which involve a small vocabulary size, their improvements tend to diminish or become \textit{marginal} when trained on more complex, large-vocabulary datasets such as \textit{WikiText-103} and \textit{One Billion Word (lm1b)}. In contrast, VQMoE consistently outperforms all competitors across benchmarks (keeping in mind that the BPC metric is log-scaled), architectures, and also converges more quickly as Figure~\ref{fig:trainingloss}. This highlights VQMoE's effectiveness in learning an efficient routing policy for the language modeling pre-training task.

\textbf{\textit{Q2: Does VQMoE keep outperforming the router method when scaling up? A2: Yes.}}

Table~\ref{table:pre-train} also demonstrates that VQMoE maintains consistently strong performance when scaled up to 12-layer Transformer and Transformer-XL architectures. Across all four datasets, the performance gap between VQMoE and other routing methods widens as the dataset size increases, from enwik8 to the One Billion Word dataset. This suggests that our approach has the potential to scale effectively with larger language models and bigger datasets. An interesting observation is that SMoE-Dropout~\citep{chen2023sparse} performs the worst among all methods, indicating that a random routing policy is insufficient and requires updating for effective training. This finding highlights that the success of SMoE-Dropout is largely due to its self-slimmable strategy, which linearly increases the number of activated experts ($K$) during training. However, this approach transforms the sparse network into a dense one, contradicting the original motivation behind using SMoE for large-scale models.

\textbf{\textit{Q3: Can VQMoE, with only 80\% of the total parameter count, achieve better performance than SMoE utilizing the full 100\% of parameters? A3: Yes.}}

\begin{figure*}[t]
    \centering
    \begin{subfigure}{.49\textwidth}
         \centering
         \includegraphics[width=\textwidth]{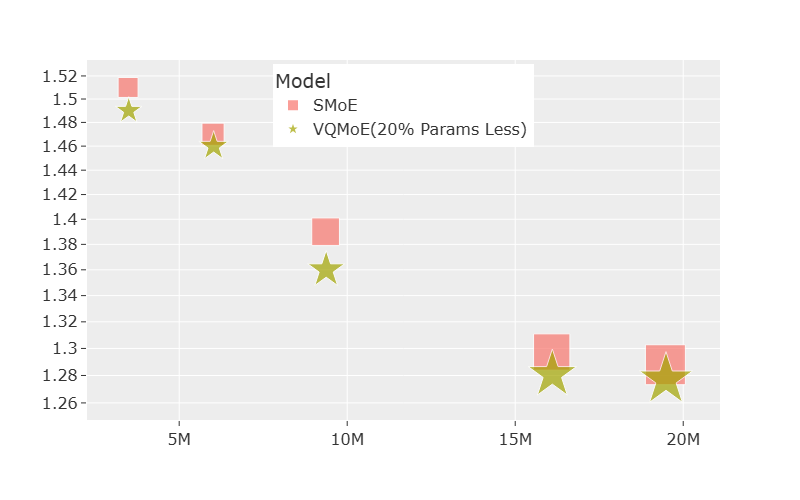}
         \caption{Robust VQMoE Benchmark (Enwik8)}
         \label{fig:ren8}
     \end{subfigure}
     \hfill
    \begin{subfigure}{.49\textwidth}
         \centering
         \includegraphics[width=\textwidth]{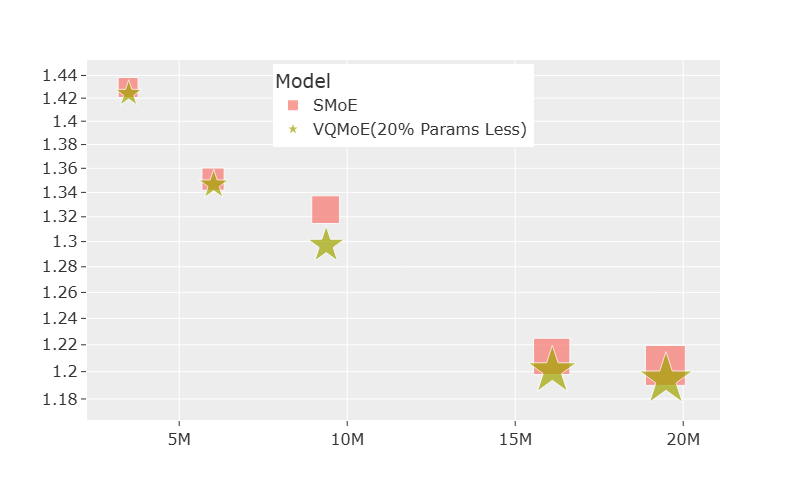}
         \caption{Robust VQMoE Benchmark (Text8)}
         \label{fig:rtx8}
     \end{subfigure}
     \hfill
     
     \caption{Illustration of the proposed Robust VQMoE architecture for Pre-training on Enwik8 and Text8 dataset. (a) Robust VQMoE architecture achieves the same performance with the routing methods while only using 80\% of the parameters on Enwik8 dataset. (b) Roubust VQMoE demonstrates robustness on the Text8 dataset. Bits-per-character (BPC) on the Enwik8 and Text8 datasets, and lower is better.} \label{fig:robust}
     \vspace{-0.1in}
\end{figure*}

To evaluate the robustness of VQMoE, we reduce its hidden dimension to half that of the SMoE baseline, resulting in approximately a 20\% reduction in the total number of parameters. Robustness here denotes the model’s ability to maintain strong performance across different parameter scales, particularly with fewer parameters. We then train both models across a range of parameter scales: 1M, 2M, 4M, 8M, and 16M, where M denotes the number of parameters in millions. Despite having only 80\% of the parameter count, VQMoE consistently achieves competitive performance compared to SMoE across all scales. This highlights the efficiency and robustness of our approach. The results are illustrated in Figure~\ref{fig:ren8} and Figure~\ref{fig:rtx8}, which show VQMoE’s performance on the Enwik8 and Text8 datasets, respectively.

\vspace{-0.1in}
\subsection{Parameter-Efficient Fine-Tuning}\label{sec:finetune}

\textbf{\textit{Q4: What is the biggest advantage of VQMoE, compared to the conventional SMoE? A4: Parameter-Efficient Fine-Tuning.}}

\begin{table}[t]
\resizebox{\linewidth}{!}{%
\begin{tabular}{@{}lccccccccc>{\columncolor{LightBlue}}c@{}}
\toprule
Architecture  & FLOPs(x$10^{10}$)              & \multicolumn{4}{c}{Transformer} & \multicolumn{4}{c}{Transformer-XL} & Avg. \\ \midrule
Dataset     &       & SST-2 & SST-5 & IMDB & BANKING77 & SST-2 & SST-5 & IMDB & BANKING77 & - \\ \midrule
VQMoE & \textbf{5.6145} & \textbf{82.6} & \textbf{41.1} & \textbf{89.5} & \textbf{84.8} & \textbf{83.3} & \textbf{42.0} & \textbf{89.1} & \textbf{85.3} & \textbf{74.72} \\
SMoE & 7.7620 & 82.1 & 39.5 & 89.3 & 82.6 & 80.8 & 40.4 & 88.6 & 80.2 & 72.94 \\
SMoE-Dropout & 7.7620 & 81.3 & 39.6 & 88.9 & 77.9 & 81.8 & 40.0 & 89.1 & 77.3 & 72.00 \\
XMoE & 7.7620 & 82.4 & 39.9 & 89.0 & 83.1 & 81.3 & 40.3 & 88.7 & 82.7 & 73.43 \\
StableMoE & 7.7620 & 82.2 & 40.4 & 89.1 & 82.7 & 82.5 & 41.1 & 88.5 & 78.6 & 73.89 \\
\bottomrule
\end{tabular}}
\caption{Accuracy of the model after fine-tuning on various datasets. Higher is better; best results are in bold.}
\label{table:finetune}
\end{table}

We see that the discrete representation that VQMoE learns at the Pretraning stage ~\ref{sec:pretrained} might consist of rich knowledge. To test this hypothesis, we use only the discrete representation for downstream tasks, allowing VQMoE to \textbf{save 28\%} of computational resources compared to SMoE. Table~\ref{table:finetune} reports the accuracy of the models fine-tuned on the test sets of various datasets. Overall, we observe that VQMoE demonstrates strong transfer learning capabilities by achieving the highest accuracy on all datasets. Notably, on the more challenging datasets of SST-5 and BANKING77, which have fewer training samples or more classes, we observe larger performance gains from VQMoE versus the SMoE baseline (over $2.5\%$ improvements compared to SMoE on average). This result shows that VQMoE can learn a discrete representation that is not only good for pre-training but also exhibits strong transfer capabilities to various downstream tasks.

\vspace{-0.1in}
\subsection{Vision} \label{sec:vision}
\textbf{\textit{Q5: Can VQMoE compete with SMoE in the Vision domain? A5: Yes.}}

To make our performance comparison informative and comprehensive, we consider two kinds of baselines that are fairly comparable to VQMoE: (1) Dense Model (Vision Transformer) ~\citep{dosovitskiy2021imageworth16x16words}; (2) SoftMoE ~\citep{puigcerver2024sparsesoftmixturesexperts} - the most advanced MoE in Vision domain. We perform two configurations for training the Mixture of Experts: (1) small - \textit{10 million parameters (10M)}; (2) \textit{large - 110 million parameters (110M)}. The result at Table ~\ref{table:vision} shows that VQMoE outperforms both Vision Transformer Dense~\citep{dosovitskiy2021imageworth16x16words}, SoftMoE~\citep{puigcerver2024sparsesoftmixturesexperts}, and other routing methods such as ~\citep{dai2022stablemoe}, ~\citep{chi2022representation} on six out of eight tasks across four image classification datasets. We conduct our experiments three times on four datasets (CIFAR-10, CIFAR-100, STL-10, and SVHN) using different seeds, reporting the average results along with the standard deviation. For the large-scale dataset ImageNet-1K, we perform a single run due to resource constraints. The average performance of our method surpasses other baselines and is more stable, as indicated by the low standard deviation.

\begin{table}[!ht]
\resizebox{\linewidth}{!}{%
\begin{tabular}{@{}lcccccccccc>{\columncolor{LightBlue}}c@{}}
\toprule
{Architecture}                & \multicolumn{5}{c}{Vision Transformer (Small)} & \multicolumn{5}{c}{Vision Transformer (Large)} & Average \\ 
{$\#$ params}                & \multicolumn{5}{c}{10M} & \multicolumn{5}{c}{110M} &  \\ \midrule
Dataset            & {Cifar10} & {Cifar100} & {STL-10} & {SVHN} & {ImageNet-1K} & {Cifar10} & {Cifar100} & {STL-10} & {SVHN} & {ImageNet-1K} & - \\ \midrule

VQMoE & \textbf{89.7}$_{\pm0.4}$            & \textbf{67.3}$_{\pm0.4}$             & 66.5$_{\pm0.3}$            & \textbf{95.6}$_{\pm0.1}$   & \textbf{54.8}&\textbf{92.8}$_{\pm0.3}$             & \textbf{67.0}$_{\pm0.5}$             & 64.3$_{\pm0.5}$           & \textbf{96.0}$_{\pm0.2}$ & \textbf{71.3} & \textbf{76.5}$_{\pm0.3}$ \\
SMoE        & 88.7$_{\pm0.2}$                      & 65.4$_{\pm0.5}$                     & 66.4$_{\pm0.1}$                     & 95.4$_{\pm0.1}$ &  52.8  & 85.7$_{\pm8.5}$                      & 55.5$_{\pm2.8}$                      & 64.4$_{\pm0.2}$                     & 94.5$_{\pm0.1}$ & 71.0 & 74.0$_{\pm1.6}$ \\

XMoE        & 88.8$_{\pm0.2}$                     & 65.5$_{\pm0.5}$                      & 66.3$_{\pm0.2}$                     & 95.4$_{\pm0.1}$ & 52.5  & 87.1$_{\pm6.4}$                      & 55.9$_{\pm0.6}$                      & \textbf{64.6}$_{\pm0.3}$                     & 94.1$_{\pm0.2}$ & 70.8  & 74.2$_{\pm1.1}$ \\
StableMoE        & 88.8$_{\pm0.1}$                      & 65.5$_{\pm0.1}$                      & 66.5$_{\pm0.2}$                     & 95.4$_{\pm0.1}$  & 52.5   & 84.7$_{\pm10.5}$                     & 55.5$_{\pm1.8}$                      & 64.3$_{\pm0.6}$                     & 94.5$_{\pm0.9}$ & 70.6  & 73.8$_{\pm1.8}$ \\ 
SoftMoE        & 85.6$_{\pm0.3}$                      & 61.4$_{\pm0.3}$                      & 65.4$_{\pm0.2}$                     & 94.8$_{\pm0.1}$ & 41.6   & 80.3$_{\pm9.7}$                     & 42.9$_{\pm1.4}$                      & 63.2$_{\pm0.5}$                     & 93.5$_{\pm0.1}$ & 68.2  & 69.7$_{\pm1.6}$ \\ \midrule
ViT (Dense)   & 89.0$_{\pm0.2}$                    & 65.7$_{\pm0.3}$                     & \textbf{66.6}$_{\pm0.2}$                   & 95.6$_{\pm0.1}$  & 52.2    & 92.2$_{\pm0.3}$                     & 60.2$_{\pm2.6}$                      & 64.1$_{\pm0.5}$                     & 96.0$_{\pm0.1}$ & 71.1 & 75.3$_{\pm0.5}$ \\ \bottomrule
\end{tabular}}
\caption{Accuracy of models evaluated on vision datasets. Higher is better, the best results are in bold.} \label{table:vision}
\end{table}

\subsection{In-depth Analysis} \label{sec:analyssis}
\begin{figure*}[t]
    \centering
    \begin{subfigure}{.3\textwidth}
         \centering
         \includegraphics[width=\textwidth]{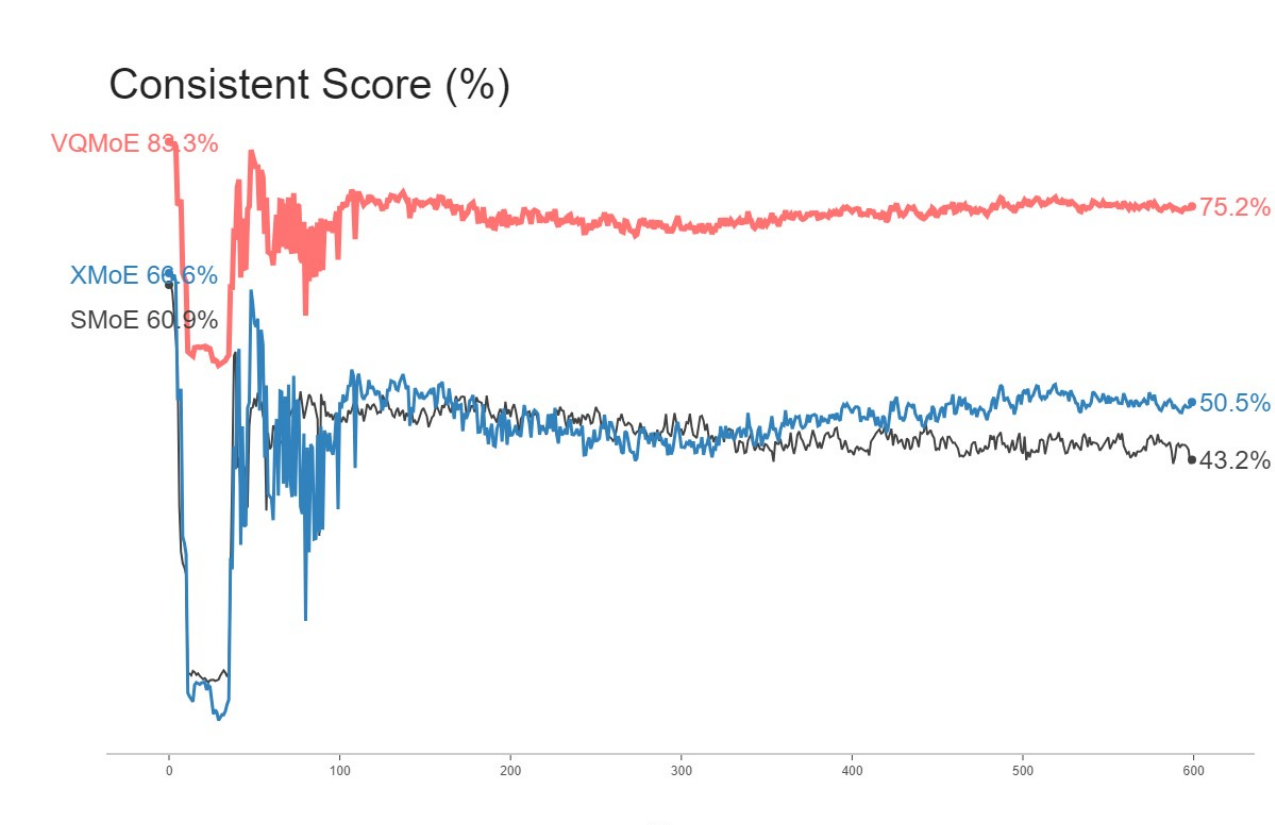}
         \caption{Consistent Score.}
         \label{fig:consist}
     \end{subfigure}
     \hfill
    \begin{subfigure}{.3\textwidth}
         \centering
         \includegraphics[width=\textwidth]{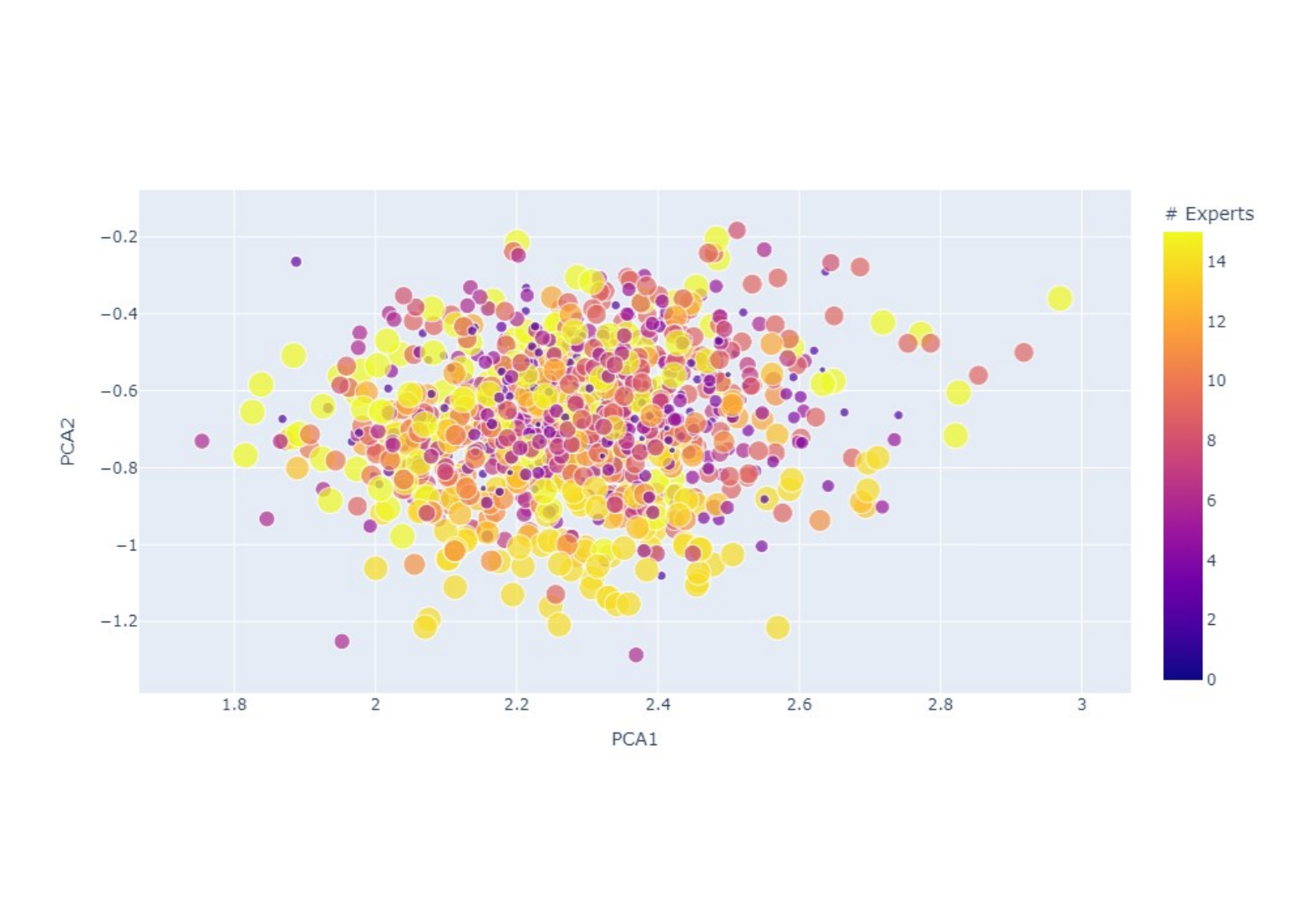}
         \caption{VQMoE Representation.}
         \label{fig:vqmoe_plot}
     \end{subfigure}
     \hfill
    \begin{subfigure}{.3\textwidth}
         \centering
         \includegraphics[width=\textwidth]{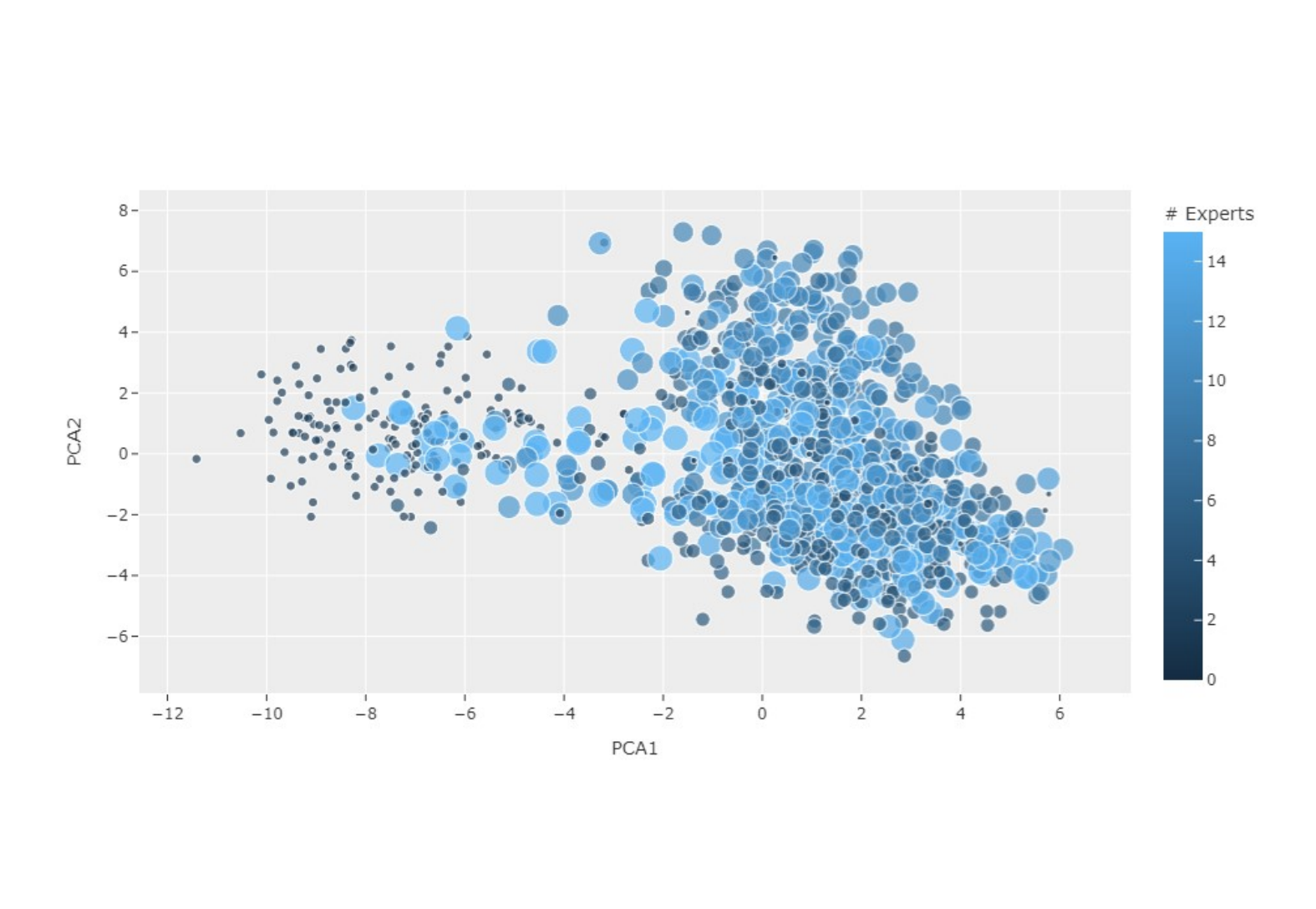}
         \caption{SMoE Representation.}
         \label{fig:smoe_plot}
     \end{subfigure}
     \hfill
     
     \caption{Analysis Inconsistent Expert Selection and Representation Collapse issues when training SMoE. Figure ~\ref{fig:consist} demonstrates consistent score movement from VQMoE, compared with SMoE and XMoE. Figure ~\ref{fig:vqmoe_plot} and Figure ~\ref{fig:smoe_plot} visualize the representation by experts in 2D dimension using Principal Component Analysis (PCA) method.}  \label{fig:analysis}
     \vspace{-0.1in}
\end{figure*}

\textbf{Consistent Score. }
Figure ~\ref{fig:consist} illustrates that expert selections when training SMoE face inconsistent problems. As the Theorem ~\ref{theorem:inconsistent_select}, this inconsistency arises because the router's coverage rate significantly exceeds that of the Transformer representation. Figure ~\ref{fig:consist} also shows that our method achieves the highest consistency score compared to the SMoE and XMoE models. However, the VQMoE model's consistency score is around 75\%, as our method also requires learning a continuous representation during the Pre-training phase.

\textbf{Representation Collapse issue. }
To visualize the Representation collapse problem in practice, we apply Principal Component Analysis (PCA) method to reduce from $d$ dimension of the Transformer to 2D for plotting purposes, thanks to ~\citep{chi2022representation}. Figures ~\ref{fig:vqmoe_plot} and ~\ref{fig:smoe_plot}  show the expert representations from the pretrained VQMoE and SMoE models. The results suggest that VQMoE experiences less representation collapse in the expert space compared to SMoE. The analysis is in line with the theorem proof at Section ~\ref{sec:theory}. However, projecting the $d$-dimensional space onto 2D for visualization may lead to information loss.

\subsection{Ablation Study} \label{sec:ablation}

We examine the effectiveness of VQMoE across various hyper-parameter settings, with all experiments conducted using the base Transformer architecture on the WikiText-103 dataset.

\textbf{Vector Quantization Method.} To learn a discrete representation, we research various types of Vector Quantization methods, including VQVAE~\citep{NIPS2017_7a98af17}, VQGAN~\citep{yu2022vectorquantizedimagemodelingimproved}, LFQ~\citep{yu2023magvitmaskedgenerativevideo}, and ResidualVQ~\citep{yang2023hificodecgroupresidualvectorquantization}. We observe that VQGAN using cosine similarity for distance achieves good and stable results in practice as Figure~\ref{fig:method}. Interestingly, VQGAN with lower dimensionality also delivers strong performance and exhibits robustness.

\textbf{Number of codebook impact.} The number of codebook entries is a crucial hyperparameter when training Vector Quantization techniques. As shown in Figure~\ref{fig:codebook}, we can see the best performance when the number of codebook entries matches the number of experts. This aligns with the proof by ~\citep{dikkala-etal-2023-benefits}, which demonstrates that in the optimal case, the number of clusters equals the number of experts.

\textbf{Sensitiveness of VQ loss contribution $\alpha$.} Figure~\ref{fig:alpha} illustrates the impact of $\alpha$, which controls the contribution of the Vector Quantization loss to the overall loss. If $\alpha$ is too high, it leads to a better discrete representation but may negatively affect the final target. Conversely, if $\alpha$ is too low, it may result in a poor discrete representation. Therefore, $\alpha$ should be selected based on the data, typically within the range of $(0.05, 0.15)$.

\section{Conclusion and Future Directions}
\label{clu}
This study illustrates Vector-Quantized Mixture of Experts (VQMoE), a novel and theoretically-grounded architecture, to overcome challenges in training SMoE such as representation collapse and inconsistency. We evaluate our method on various Pre-training and Fine-tuning tasks, for both language and vision domains. The results show that VQMoE outperforms the routing methods both theoretically and empirically. Furthermore, fine-tuning VQMoE with the discrete representation for downstream tasks could reduce computational resource usage by 28\%. We believe that focusing on discrete representation learning will offer a promising strategy for training and testing sparse mixtures of experts (SMoE) at a large scale. Finally, we believe that our approach opens up new research avenues for effectively training SMoE, where cutting-edge techniques in discrete representation learning and vector quantization can be harnessed to enhance their performance.

\bibliography{main}
\bibliographystyle{tmlr}

\appendix
\section{Appendix}
\label{add_on}
\begin{center}
{\bf{\Large{Supplementary Material for ``On the Role of Discrete Representation in Sparse Mixture of Experts''}}}
\end{center}


This document is organized as follows. Appendix~\ref{app:proof} provides a detailed proof for Section~\ref{sec:theory}. Appendix~\ref{app:add_result} presents additional experimental results demonstrating the effectiveness of our method compared to the baselines. Finally, Appendix~\ref{app:colapse_rep} offers an in-depth analysis of representation collapse, while Appendix~\ref{app:add_exp} details the implementation aspects.



\subsection{Proof for Results in Section~\ref{sec:theory}} \label{app:proof}

\subsubsection{Jacobian Matrix of VQMoE}\label{app:vqproof}

To investigate whether VQMoE alleviates this collapse, we derive the Jacobian of the VQMoE output with respect to the input \( x \in \mathbb{R}^{n \times d} \):




\begin{equation}
\label{eq:jacobian_vqmoe_v2}
\begin{aligned}
\boldsymbol{J}^{\text{VQMoE}} 
&= g_c(\boldsymbol{x}) \cdot \boldsymbol{J}^{\text{SMoE}} 
+ \frac{\partial g_c(\boldsymbol{x})}{\partial \boldsymbol{x}} f^{\mathrm{SMoE}}(\boldsymbol{x}) \\
&\quad + g_d(\boldsymbol{x}) \cdot \sum_{l=1}^K \boldsymbol{J}^{\text{FFN}}_l 
+ \frac{\partial g_d(\boldsymbol{x})}{\partial \boldsymbol{x}} \sum_{l=1}^K f_l^{\mathrm{FFN}}(\tilde{\boldsymbol{x}}_l) \\
&= g_c(\boldsymbol{x}) \cdot \left[ J_1 + \sum_{j=1}^N \boldsymbol{c}_j \boldsymbol{e}_j^\top \right]
+ \sum_{m \in \{c, d\}} g_m \boldsymbol{e}_m^\top
+ \sum_{l=1}^K d_l \boldsymbol{e}_l^\top \\
&= J_1 + \sum_{j=1}^N c_j \boldsymbol{e}_j^\top 
+ \sum_{l=1}^K d_l \boldsymbol{e}_l^\top 
+ \sum_{m \in \{c, d\}} g_m \boldsymbol{e}_m^\top \\
&= J_1 + \sum_{j=1}^{N+K+2} o_j \boldsymbol{e}_j^\top.
\end{aligned}
\end{equation}

where:


\begin{alignat*}{2}
\boldsymbol{e}_j \quad & :\quad && \text{Embedding of the } j\text{-th expert in the SMoE;} \\
J_1 = \mathcal{S}(x)_k \boldsymbol{J}^{\mathrm{FFN}} \quad & :\quad && \text{Jacobian of the top-}k \text{ FFN block;} \\
\end{alignat*}




As in SMoE, the Jacobian of VQMoE consists of two major components: \( J_1 \) - the primary contribution from the input and selected expert; and \( \sum_{i=1}^{N+K+2} o_i \boldsymbol{e}_i^\top \) - additional gradient contributions from both the continuous part and the discrete part.

\subsubsection{Proof of Theorem~\ref{theorem:inconsistent_select}}
\label{appendix:inconsistent_select}




In this proof, we use contradiction to establish the theorem. Assume that the expert embeddings \( e \) form a consistent router. By Definition~\ref{definition:consistent}, we have:  
\[
\text{dist}(x_p, u_i) \leq \min(\text{dist}(x_p, u_j)),
\]  
where \( u_i \) is the representation corresponding to the closest expert \( e_i \).

According to \citep{chi2022representation}, projecting information from a hidden representation space \( \mathbb{R}^d \) to the expert dimension \( N \) leads to representation collapse. Now, consider three output of Multi-Head Attention (MHA) (MHA) layer: \( x_1, x_2, x_3 \) $\in \mathbb{R}^d$, belong to experts whose embeddings \( e_1, e_2, e_3 \) collapse. Without loss of generality, assume that \( e_2 \) lies between \( e_1 \) and \( e_3 \) in the embedding space. Then, we have:  

\begin{equation}
\begin{aligned}
\text{dist}(x_2, u_2) &\leq \min(\text{dist}(x_1, e_1), \text{dist}(x_2, e_2), \text{dist}(x_3, e_3)) \\
&\leq \text{dist}(e_1, e_3).
\end{aligned}
\end{equation}

Let \( t_e \) denote the step at which the embeddings \( e_1 \) and \( e_3 \) converge, and \( t_m \) denote the step at which the Multi-Head Attention (MHA) module converges. From step \( t_e \), it follows that:  
\[
\lim_{t_e \to t_m} \text{dist}(x_2, u_2) = \lim_{t_e \to t_m} \text{dist}(e_1, e_3) = 0.
\]

Thus, \( y \) (the output of MHA) converges at step \( t_e \).  

This directly contradicts the assumption that the MHA converges at step \( t_m \), where \( t_e \ll t_m \).

\subsubsection{Proof of Proposition~\ref{prop:exp_discrete}}
\label{appendix:optimal}





We use contradiction to prove the proposition. Assume that, at training step \( t \), there exists a set of pairs \( (C_i, E_j) \) such that \( i \neq j \). Let \( x_1, x_2, \ldots, x_N \) represent a sequence of inputs sampled from \( N \) clusters. From step \( t_0 \) to step \( t_{m-1} \), each pair \( (x_j, E_j) \), where \( j \in [1, N] \), is updated using the following gradient descent equation:

\[
W^{t_{l+1}}_{E_j} = W^{t_l}_{E_j} - \eta \mathcal{J}(x_j),
\]

where \( W^{t_l}_{E_j} \) is the weight of expert \( E_j \) at iteration \( t_l \), \( \mathcal{J}(x_j) \) is the Jacobian matrix with respect to input \( x_j \), and \( \eta \) is the learning rate, and $0 \leq l<m-1$.

Let \( \mathcal{L} \) denote the loss function during the training process described by Equation~\ref{eqa:loss}. After \( t_{m-1} \) training steps, the following condition holds:   
\[
\mathcal{L} (E_j(x_j)) = \min_{c \in [1,N]} \mathcal{L}(E_c(x_j)).
\]

Under the assumption of contradiction, there exists a set of pairs, where $x_j$ is assigned to an expert $E_i$: $(x_j, E_i)$ ; $i, j \in[1,N]$ and $i \neq j$; where the loss function \( \mathcal{L} \) is minimized. It means:
\[
\mathcal{L} (E_i(x_j)) \leq \mathcal{L}(E_j(x_j))
\]  
However, by definition of the loss minimization process, the inequality  
\[
\mathcal{L} (E_j(x_j)) \leq \mathcal{L}(E_i(x_j))
\]  
must hold.

This leads to a contradiction with our initial assumption. 

\subsection{Additional Experiment Results}\label{app:add_result}

\textbf{\textit{Q6: Can VQMoE learn Discrete Representation Only from scratch? A6: Yes for small scale, but no for large scale.}}

The answer is yes for small models. However, training a discrete representation-only approach is feasible primarily for small-scale models with a moderately sized dataset. The results of the \textit{Transformer-XL} model in Table~\ref{tab:moe_comparison} on the Enwik8 dataset support this observation. As the model scales up, relying solely on discrete representation reaches its limitations, leading to performance below the SMoE baselines.

\begin{table}[t]
    \centering
    \begin{tabular}{l c c c c}
        \toprule
        Scale & TopK & \# Experts & SMoE & VQMoE (Discrete Only) \\
        \midrule
        \multirow{5}{*}{Base 20M-50K Steps}  & 1  & 16 & 1.28 & \textbf{1.25} \\
          & 2  & 16 & 1.26 & - \\
          & 4  & 16 & 1.26 & - \\
          & 8  & 16 & 1.27 & - \\
          & 16 & 16 & 1.27 & - \\ \hline
        \multirow{5}{*}{Base 20M-100K Steps}   & 1  & 16 & 1.22 & \textbf{1.18} \\
          & 2  & 16 & 1.20 & - \\
          & 4  & 16 & 1.21 & - \\
          & 8  & 16 & 1.21 & - \\
          & 16 & 16 & 1.21 & - \\ \hline
        \multirow{7}{*}{Large (210M)} & 1  & 64 & \textbf{1.12} & 1.14 \\
         & 2  & 64 & 1.09 & - \\
         & 4  & 64 & 1.09 & - \\
         & 8  & 64 & 1.09 & - \\
         & 16 & 64 & 1.10 & - \\
         & 32 & 64 & 1.10 & - \\
         & 64 & 64 & 1.12 & - \\
        \bottomrule
    \end{tabular}
    \caption{Performance comparison of SMoE and VQMoE (Discrete Only) on the \textit{Enwik8} (BPC) dataset.}
    \label{tab:moe_comparison}
\end{table}

\textbf{\textit{Q7: Can VQMoE outperform the clustering-based approach such as KMean? A7: Yes.}}


We explored a clustering-based approach -MoCLE\citep{gou2024mixtureclusterconditionalloraexperts},  but found it unsuitable for our method. Unlike MoCLE, Vector Quantization allows the model greater flexibility in learning cluster representations during training, making it more competitive in practical applications. The training results using the Transformer-XL model on the Enwik8 dataset are presented in Table~\ref{tab:cluster_comparison}.

\begin{table}[t]
    \centering
    \begin{tabular}{l c c c c c}
        \toprule
        Scale & TopK & \# Experts & SMoE & MoCLE & VQMoE  \\
        \midrule
         \multirow{5}{*}{Base 20M-50K Steps} & 1  & 16 & 1.28 & 1.29 & \textbf{1.25}  \\
         & 2  & 16 & 1.26 & 1.28 & -   \\
        & 4  & 16 & 1.26 & 1.28 & -   \\
        & 8  & 16 & 1.27 & 1.28 & -   \\
        & 16 & 16 & 1.27 & 1.28 & -   \\
        \bottomrule
    \end{tabular}
    \caption{Performance comparison of VQMoE and MoCLE (Clustering approach) on the \textit{Enwik8} (BPC) dataset.}
    \label{tab:cluster_comparison}
\end{table}

\textbf{\textit{Q8: Can VQMoE contribute to AI real-world applications? A8: Yes.}}

We found that VQMoE can directly benefit real-world AI applications, such as image segmentation, demonstrating its strong generalization capabilities. Specifically, our method outperforms both the baseline and dense models in terms of Mean Accuracy and mIoU metrics on the ADE20K dataset~\citep{zhou2018semanticunderstandingscenesade20k} using the Segmenter model\citep{strudel2021segmentertransformersemanticsegmentation}. Detailed results are provided in Table~\ref{tab:segmentation_comparison}.


\begin{table}[t]
    \centering
    \begin{tabular}{l c c c c c c c}
        \toprule
        Model & ViT & SoftMoe & SMoE & StableMoE & XMoE & VQMoE & Metrics \\
        \midrule
        \multirow{2}{*}{Segmenter} & 20.8 & 19.0 & 23.1 & 22.4 & 22.3 & \textbf{23.4} & Mean accuracy \\
         & 15.0 & 14.0 & 15.5 & 16.0 & 15.7 & \textbf{16.6} & mIoU \\
        \bottomrule
    \end{tabular}
    \caption{Comparison of VQMoE versus the baselines on the ADE20K dataset.}
    \label{tab:segmentation_comparison}
\end{table}

\textbf{\textit{Q9: Does VQMoE consistently outperform the baselines across multiple training runs? A9: Yes.}}

Due to resource constraints, it is challenging to train all models across all datasets multiple times and to perform formal statistical significance testing. To illustrate the variance across multiple training runs, we train VQMoE and baseline models on the Text8 dataset three times. The average Bits-Per-Character (BPC) and standard deviation for each model are reported in Table~\ref{tab:repeated_runs}. The results indicate that VQMoE achieves the best average performance, while also exhibiting a lower standard deviation compared to other models, suggesting greater training stability. The consistency observed across repeated runs supports the reliability of the results reported in Table~\ref{tab:repeated_runs}.

\begin{table}[h]
\centering
\caption{Average BPC and standard deviation across three training runs on Text8. Lower is better; best results are in bold.}
\label{tab:repeated_runs}
\begin{tabular}{@{}lllc@{}}
\toprule
\textbf{Model} & \textbf{Algorithm} & \textbf{Dataset} & \textbf{Avg. BPC} \\
\midrule
Transformer-XL & VQMoE     & Text8 & $\textbf{1.280} \pm 0.003$ \\
               & SMoE      &       & $1.293 \pm 0.007$ \\
               & XMoE      &       & $1.282 \pm 0.003$ \\
               & StableMoE &       & $1.285 \pm 0.005$ \\
\bottomrule
\end{tabular}
\end{table}

\textbf{\textit{Q10: Is VQMoE able to consistently surpass SMoE models in large-scale evaluation scenarios? A10: Yes.}}

We explore a more extensive model variant, OLMoE-1B-7B~\cite{muennighoff2025olmoeopenmixtureofexpertslanguage}, which comprises 16 layers, 7 billion parameters, 64 experts, and a top-$k$ selection of 8. Due to limitations in time and computational resources, we utilize the pre-trained routers for codebook embedding and compare our proposed VQMoE with OLMoE in a training-free setting. The evaluation is conducted across 6 diverse tasks and 19 datasets from the Massive Text Embedding Benchmark (MTEB)~\cite{muennighoff2022mteb}. The summary of this evaluation is provided in Table~\ref{tab:olmoe-vqmoe}.

\begin{table}[h!]
\centering
\caption{Zero-shot performance comparison between OLMoE and VQMoE on MTEB. The best score per dataset is highlighted in bold. Improvement (Imp.) is calculated as \texttt{(GAP / OLMoE) * 100}}
\label{tab:olmoe-vqmoe}
\small 
\begin{tabularx}{\textwidth}{llcccccX}
\toprule
\textbf{Task} & \textbf{Dataset} & \textbf{Params} & \textbf{\#Exp} & \textbf{Top-$k$} & \textbf{OLMoE} & \textbf{VQMoE} & \textbf{Imp. (\%)} \\
\midrule
\multirow{3}{*}{Classification} 
  & Emotion & 7B & 64 & 8 & 49.9 & \best{52.5} & 5.2 \\
  & Toxic & 7B & 64 & 8 & 65.2 & \best{67.2} & 3.1 \\
  & Tweet & 7B & 64 & 8 & 58.0 & \best{59.8} & 3.1 \\
\midrule
\multirow{2}{*}{Clustering} 
  & Medrxiv & 7B & 64 & 8 & 23.9 & \best{25.8} & 7.5 \\
  & 20Groups & 7B & 64 & 8 & 25.7 & \best{28.4} & 10.5 \\
\midrule
\multirow{2}{*}{Pair Classification} 
  & SemEval & 7B & 64 & 8 & 46.7 & \best{49.5} & 6.0 \\
  & URLCorpus & 7B & 64 & 8 & 77.4 & \best{79.4} & 2.6 \\
\midrule
\multirow{3}{*}{Reranking} 
  & Ask & 7B & 64 & 8 & 51.9 & \best{53.3} & 2.7 \\
  & SciDocs & 7B & 64 & 8 & 69.6 & \best{72.3} & 3.7 \\
  & StackOver & 7B & 64 & 8 & 32.5 & \best{33.9} & 4.3 \\
\midrule
\multirow{7}{*}{STS} 
  & Biosses & 7B & 64 & 8 & 61.8 & \best{68.7} & 11.2 \\
  & SickR & 7B & 64 & 8 & 65.7 & \best{66.5} & 1.4 \\
  & STS12 & 7B & 64 & 8 & 53.8 & \best{56.0} & 4.1 \\
  & STS13 & 7B & 64 & 8 & 66.5 & \best{74.0} & 11.3 \\
  & STS14 & 7B & 64 & 8 & 56.8 & \best{59.5} & 4.6 \\
  & STS15 & 7B & 64 & 8 & 69.3 & \best{71.5} & 3.2 \\
  & STS16 & 7B & 64 & 8 & 70.1 & \best{70.5} & 0.6 \\
\midrule
Summarization & Medrxiv & 7B & 64 & 8 & 28.9 & \best{29.8} & 3.1 \\
\midrule
\textbf{Average} & -- & -- & -- & -- & 54.1 & \best{56.6} & 4.6 \\
\bottomrule
\end{tabularx}
\end{table}

Interestingly, we find that VQMoE consistently outperforms OLMoE across all tasks and datasets, despite not undergoing additional training or fine-tuning. On average, across six tasks, VQMoE shows a relative improvement of 4.6\%. The most significant gains appear in the Classification and Clustering tasks. These findings support our hypothesis that VQMoE enhances pre-trained models by learning more effective routing policies. Furthermore, by mitigating representation collapse through the use of discrete representations, VQMoE improves the model’s overall representational capacity.

\subsection{Representation Collapse Analysis}\label{app:colapse_rep}
\begin{figure*}[t]
    \centering
    \begin{subfigure}{.48\textwidth}
         \centering
         \includegraphics[width=\textwidth]{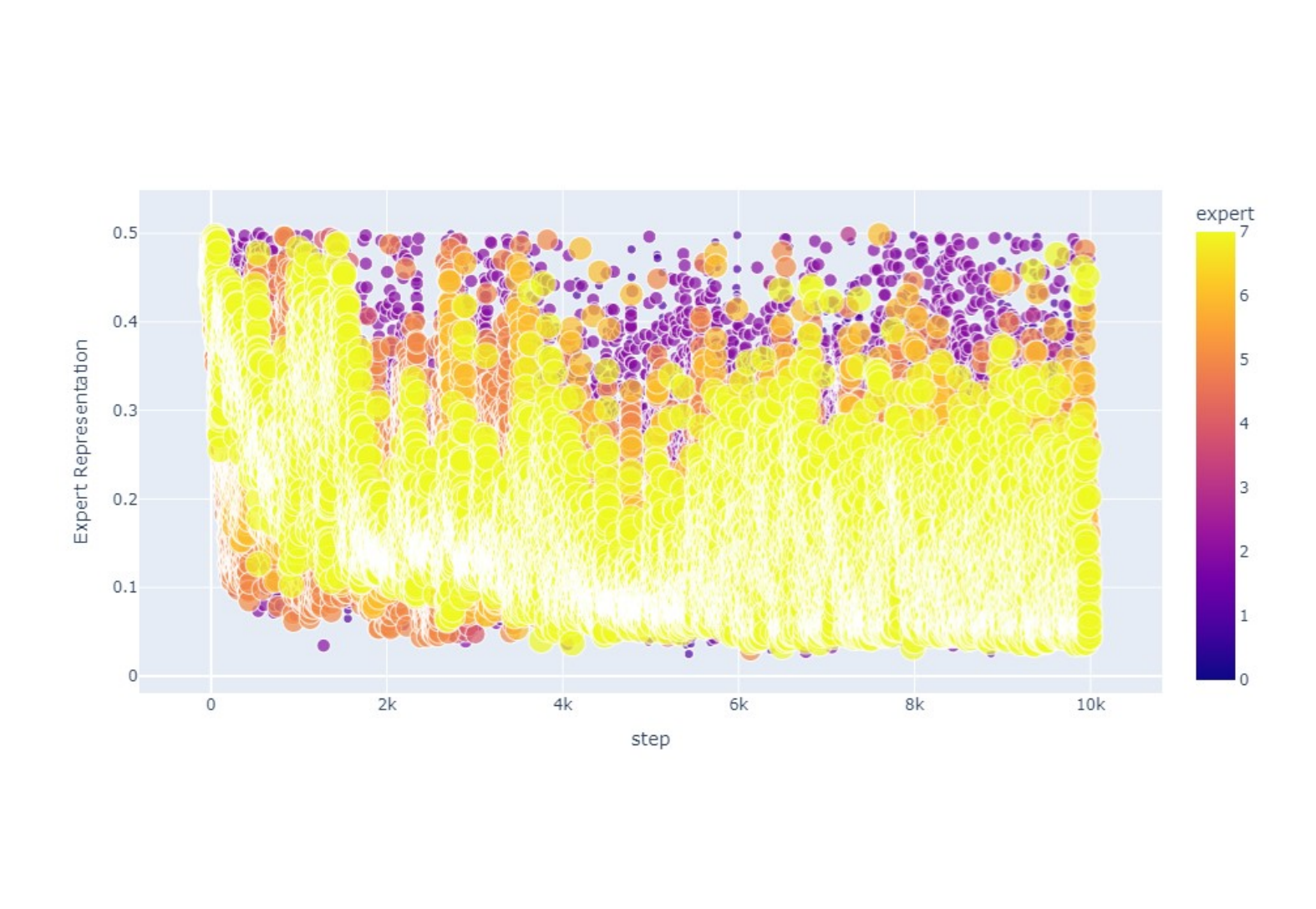}
         \caption{Training Input Token Representations. } 
         \label{fig:exp_repre}
     \end{subfigure}
     \hfill
    \begin{subfigure}{.48\textwidth}
         \centering
         \includegraphics[width=\textwidth]{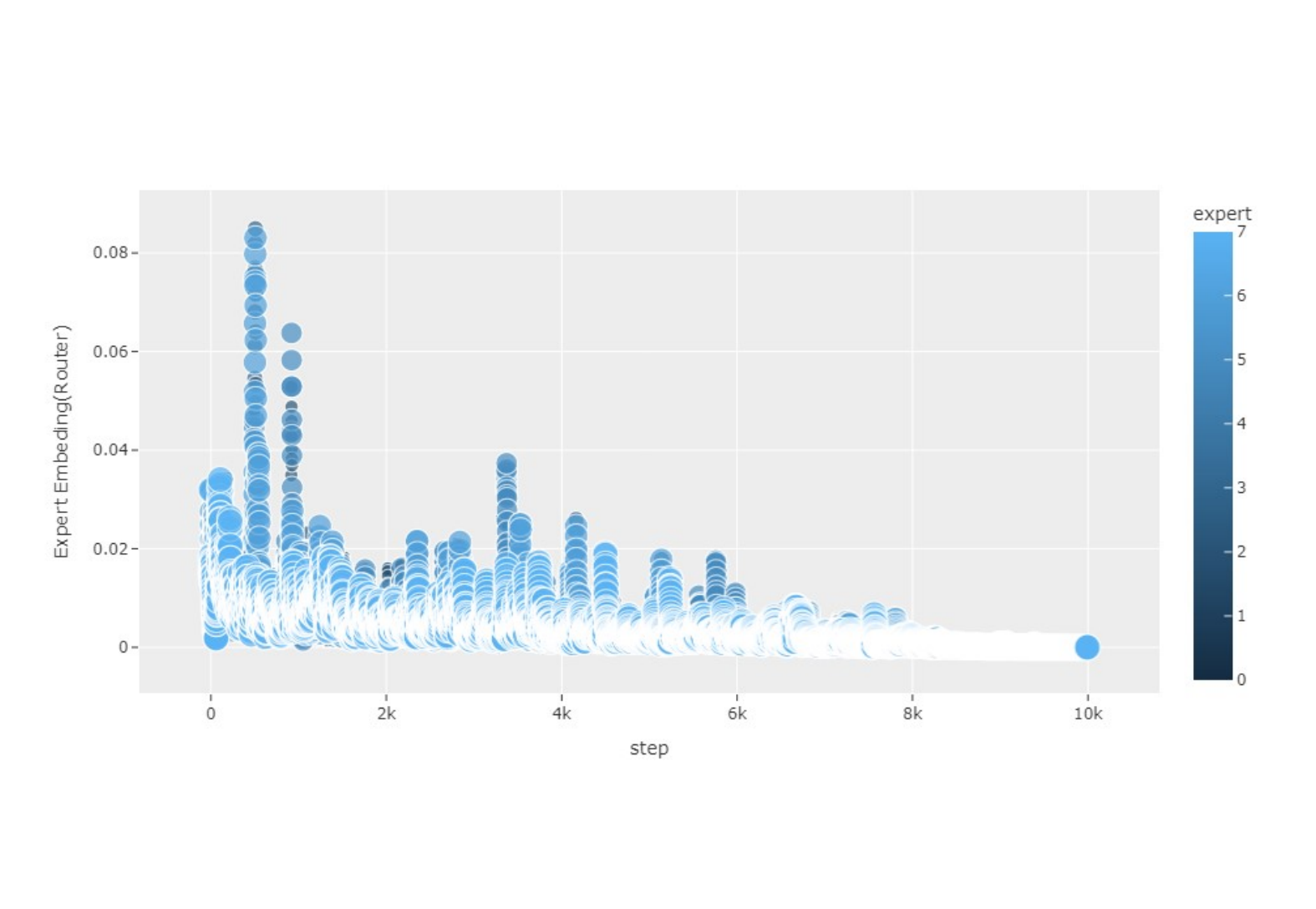}
         \caption{Training Router Representation (Expert embedding).} 
         \label{fig:router_emb}
     \end{subfigure}
     \caption{Comparison of Token Representation and Expert Representation across Training Iteration.} \label{fig:rep}
     \hfill
     \vspace{-0.1in}
\end{figure*}

To illustrate Theorem ~\ref{theorem:inconsistent_select}, we perform a language model task as described in Section ~\ref{sec:setting}, examining the movement of Expert Input Representation in Figure ~\ref{fig:exp_repre} and Expert Embedding (router) in Figure ~\ref{fig:router_emb}. We analyze the dynamics of the expert input representations by tracking their changes across training iterations. The results indicate that the inputs to the experts become increasingly divergent over time. This divergence suggests that the model learns to represent the data in a more specialized and diverse manner, allowing each expert to focus on distinct features or patterns within the data. Similarly, we track the changes in expert embeddings (router) throughout the training process. However, the trend is the opposite: the expert embeddings appear to converge quickly, stabilizing around 10,000 iterations. The findings align with our assumption stated in Theorem ~\ref{theorem:inconsistent_select}, indicating that Expert Embedding converges more quickly than Expert Input Representation. These results provide further evidence supporting the Theorem ~\ref{theorem:inconsistent_select}.

     

     

\subsection{Experiments implementation details} \label{app:add_exp}

This section provides detailed parameters of our experiments in Section \ref{sec:exp}. 

\subsubsection{General Settings} \label{app:setting}
The experiments are based on the publicly available SMoE-Dropout implementation\citep{chen2023sparse}\footnote{\url{https://github.com/VITA-Group/Random-MoE-as-Dropout}}. However, the pre-training was conducted on two H100 GPUs, so results might differ when using parallel training on multiple GPUs.

\subsubsection{Pre-training Experiments}\label{sec:setting}
Table \ref{tab:A1} provides the detailed configurations for pre-training Transformer~\citep{Vaswani+2017}, Transformer-XL~\cite{dai2019transformerxl} on \texttt{Enwik8}, \texttt{Text8}, \texttt{WikiText-103},and \texttt{One Billion Word}.


\begin{table}[!ht]
\centering
\setlength\tabcolsep{3.06pt}

\begin{tabular}{lccccccc}
\midrule
Dataset   & Input length & Batch size & Optimizer & Lr   & \# Training Step & \# Experts & TopK \\ \midrule
\texttt{Enwik8}      & 512          & 48          & Adam      & 3.5e-4 & 100k   & 16 & 2    \\ 
\texttt{Text}      & 512          & 48          & Adam      & 3.5e-4 & 100k    & 16 & 2          \\ 
\texttt{WikiText-103}      & 512          & 22          & Adam      & 3.5e-4 & 100k  & 16 & 2            \\ 
\texttt{One Billion Word}      & 512          & 11          & Adam      & 3.5e-4 & 100k    & 16 & 2          \\ \midrule
\end{tabular}
\caption{Hyperparameter settings for pre-training experiments on \texttt{Enwik8}, \texttt{Text8} , \texttt{WikiText-103} , and \texttt{One Billion Word}. }
\label{tab:A1}
\end{table}
\begin{table}[!ht]
\centering

\setlength\tabcolsep{4.86pt}
\begin{tabular}{lccccc}
\midrule
Dataset   & Input length & Batch size & Optimizer & Lr   & \# Epochs \\ \midrule
\texttt{SST-2}     & 512          & 16         & Adam      & 1e-4 & 5         \\
\texttt{SST-5}     & 512          & 16         & Adam      & 1e-4 & 5         \\
\texttt{IMDB}      & 512          & 4          & Adam      & 1e-4 & 5         \\
\texttt{BANKING77} & 512          & 16         & Adam      & 1e-4 & 5         \\ \midrule
\end{tabular}
\caption{Detail settings for fine-tuning experiments on the evaluation datasets. }
\label{tab:A2}
\end{table}

\subsubsection{Fine-tuning Experiments}
\noindent For fine-tuning experiments, we employ the identical model architecture as in pre-training. Table \ref{tab:A2} presents the detailed configurations utilized for fine-tuning experiments on \texttt{SST-2}, \texttt{SST-5}, \texttt{IMDB}, and \texttt{BANKING77} datasets.
We start with the pretrained checkpoint of the base model on enwik8, remove the final layer, and replace it with two randomly initialized fully connected layers to serve as the classifier for each fine-tuning dataset. All methods are fine-tuned for 5,000 steps with a uniform learning rate. 

\begin{figure*}[t]
    \centering
    \begin{subfigure}{.3\textwidth}
         \centering
         \includegraphics[width=\textwidth]{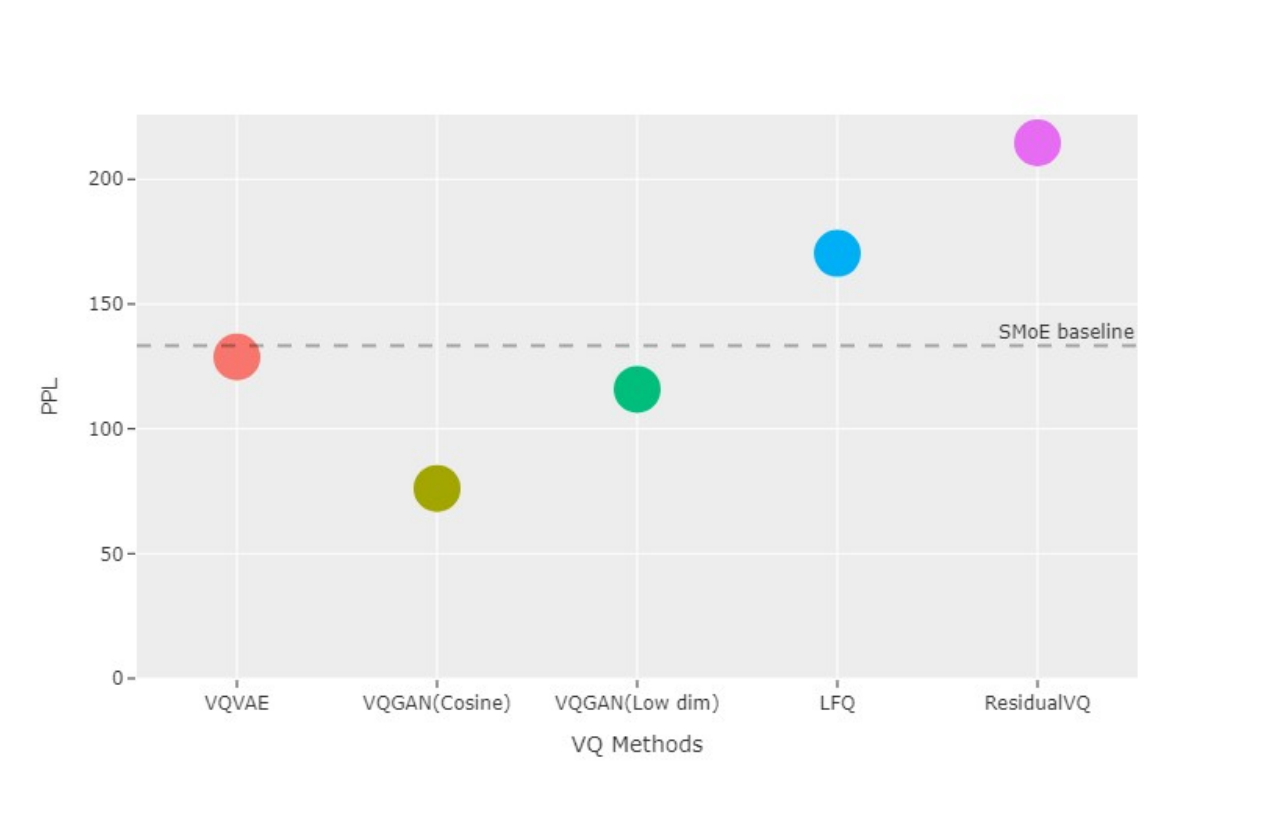}
         \caption{Vector Quantization method.}
         \label{fig:method}
     \end{subfigure}
     \hfill
    \begin{subfigure}{.3\textwidth}
         \centering
         \includegraphics[width=\textwidth]{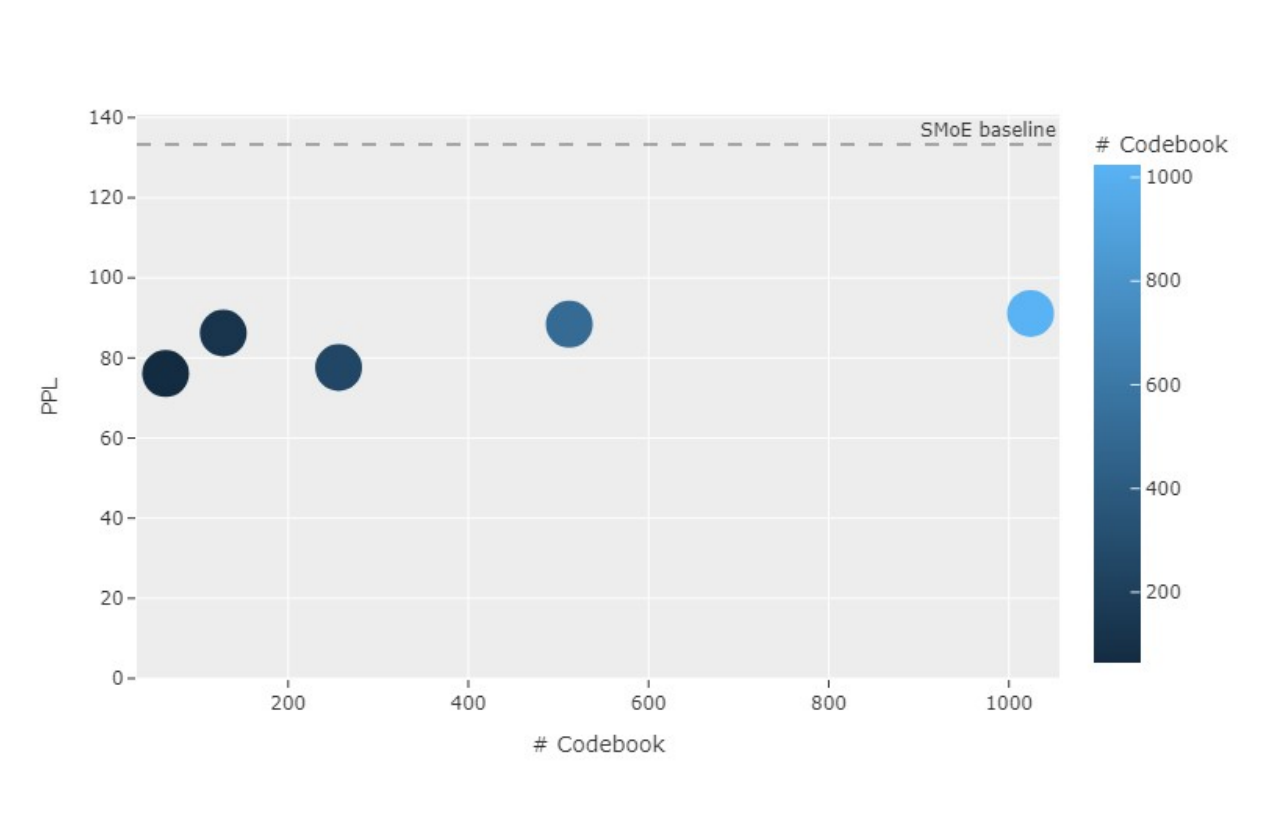}
         \caption{Number of codebook.}
         \label{fig:codebook}
     \end{subfigure}
     \hfill
    \begin{subfigure}{.3\textwidth}
         \centering
         \includegraphics[width=\textwidth]{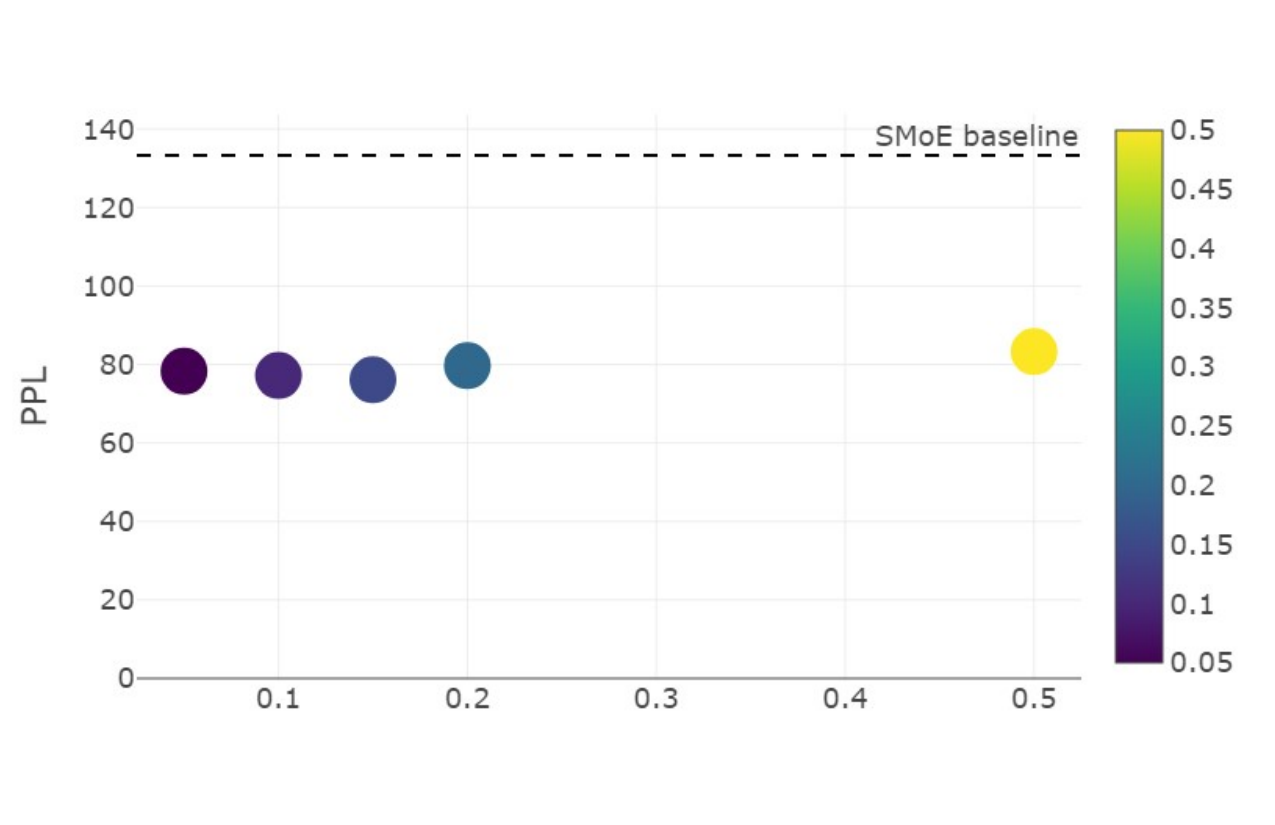}
         \caption{Impact of $
         \alpha$ for VQMoE.}
         \label{fig:alpha}
     \end{subfigure}
     \hfill
     
     \caption{Pre-training small Transformer-XL on WikiText-103 across different hyperparameters.} \label{fig:tuning}
     \vspace{-0.1in}
\end{figure*}


\end{document}